\documentclass[10pt,twocolumn,letterpaper]{article}

\usepackage{cvpr}
\usepackage{times}
\usepackage{epsfig}
\usepackage{graphicx}
\usepackage{amsmath}
\usepackage{amssymb}
\usepackage{soul}

\usepackage{graphics}
\usepackage[usenames,dvipsnames,table]{xcolor}
\usepackage{array}
\usepackage{multirow}
\usepackage{wrapfig}
\usepackage{hhline}
\usepackage{pifont}
\newcommand{\showimagew}[2][\linewidth]{\includegraphics[width={#1}]{{#2}}}

\DeclareMathOperator*{\argmin}{arg\,min} 
\newcommand{\ve}{\boldsymbol} 
\newcommand{\refeq}[1]{(\ref{#1})} %
\newcommand{\reffig}[1]{Figure~\ref{#1}} 

\newcommand{\topscore}[1]{\textcolor{blue}{\textbf{#1}}} 
\newcommand{\failcase}[1]{\textcolor{red}{\textbf{#1}}} 
\newcolumntype{K}[1]{>{\centering\arraybackslash}p{#1}}
\newcommand{\negvspace}{\vspace{-0.2cm}}

\usepackage{caption}

\usepackage[symbol]{footmisc}

\usepackage[pagebackref=true,breaklinks=true,letterpaper=true,colorlinks,bookmarks=false]{hyperref}

\cvprfinalcopy 

\def\cvprPaperID{23} 

\ifcvprfinal\fi
\begin{document}

\title{Information-Flow Matting\footnote[2]{alo naber}
\negvspace
\negvspace}

\author{Ya\u{g}{\i}z Aksoy$^{1,2}$, Tun\c{c} Ozan Ayd{\i}n$^{2}$, Marc Pollefeys$^{1}$\negvspace\\\\\negvspace
$^{1}$ ETH Z\"{u}rich \quad\quad $^{2}$ Disney Research Z\"{u}rich
}

\twocolumn[{%
\renewcommand\twocolumn[1][]{#1}%
\maketitle
\begin{center}
    \centering
    \negvspace
    \negvspace
    \showimagew[\linewidth]{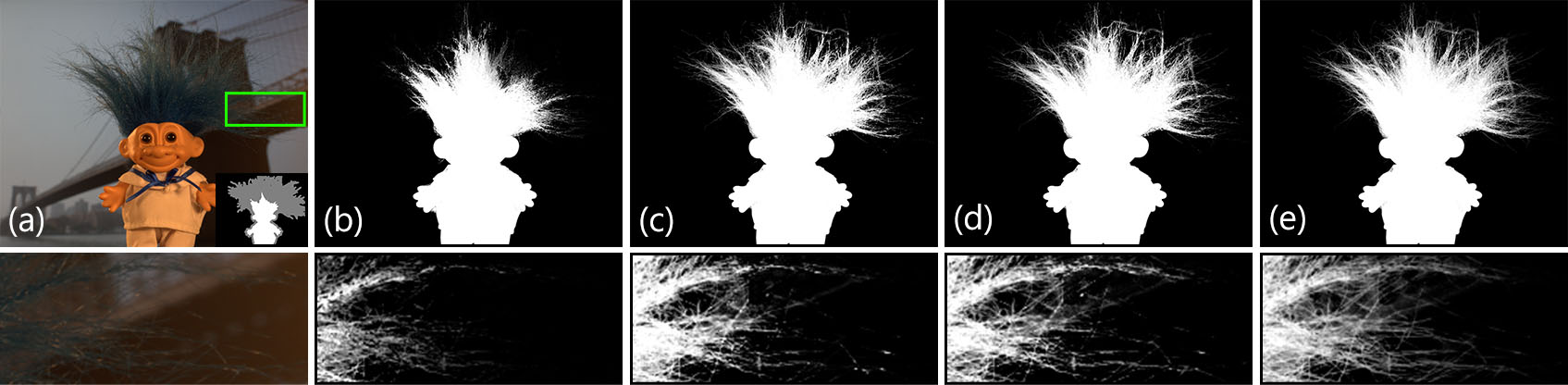}
    \negvspace
    \negvspace
    \captionof{figure}{
        For an input image and a trimap (a), we construct our linear system by first using the color-mixture flow(b), then adding direct channels of information flow from known to unknown regions (c), and letting information be shared effectively inside the unknown region (d).
        We finally introduce local information flow to enforce spatial smoothness (e).
        Note that the intermediate results in this figure are solely for illustration.
        In practice, we construct a single energy function that accounts for all types of information flow and solve it once to obtain the end result.
    }
    \negvspace
    \label{fig:teaser}
\end{center}%
}]

\maketitle

\negvspace
\negvspace
\footnotetext[2]{This document is an extended version of the 2017 CVPR publication titled \emph{Designing effective inter-pixel information flow for natural image matting}.}
\begin{abstract}
    \vspace{-0.3cm}
We present a novel, purely affinity-based natural image matting algorithm. Our method relies on carefully defined pixel-to-pixel connections that enable effective use of information available in the image. We control the information flow from the known-opacity regions into the unknown region, as well as within the unknown region itself, by utilizing multiple definitions of pixel affinities. Among other forms of information flow, we introduce color-mixture flow, which builds upon local linear embedding and effectively encapsulates the relation between different pixel opacities. Our resulting novel linear system formulation can be solved in closed-form and is robust against several fundamental challenges of natural matting such as holes and remote intricate structures. 
While our method is primarily designed as a standalone matting tool, we show that it can also be used for regularizing mattes obtained by sampling-based methods. The formulation is also extended to layer color estimation and we show that the use of multiple channels of flow increases the layer color quality. We also demonstrate our performance in green-screen keying and analyze the characteristics of the utilized affinities.
\negvspace
\negvspace
\end{abstract}

\section{Introduction}
\label{sec:intro}

Extracting the opacity information of foreground objects from an image is known as natural image matting.
Natural image matting has received great interest from the research community through the last decade and can nowadays be considered as one of the classical research problems in visual computing.
Mathematically, image matting requires expressing pixel colors in the transition regions from foreground to background as a convex combination of their underlying foreground and background colors.
The weight, or the \emph{opacity}, of the foreground color is referred to as the alpha value of that pixel.
Since neither the foreground and background colors nor the opacities are known, estimating the opacity values is a highly ill-posed problem.
To alleviate the difficulty of this problem, typically a \emph{trimap} is provided in addition to the original image.
The trimap is a rough segmentation of the input image into foreground, background, and regions with unknown opacity.

\begin{figure*}
{
\footnotesize
\begin{tabular}
{   K{0.087\linewidth}
    K{0.087\linewidth}
    K{0.087\linewidth}
    K{0.087\linewidth}
    K{0.087\linewidth}
    K{0.087\linewidth}
    K{0.087\linewidth}
    K{0.087\linewidth}
    K{0.087\linewidth}
}
Input&
Gnd-truth&
Trimap&
Closed-form&
KNN - HSV&
KNN - RGB&
Man. Pres.&
CMF-only&
Ours
\end{tabular}
}
\centering
\showimagew[\linewidth]{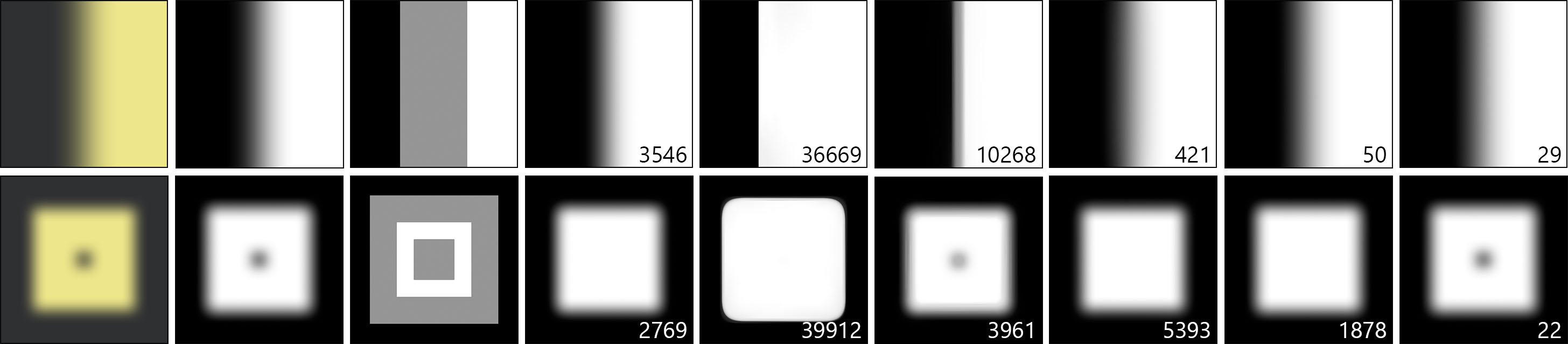}
\caption{
We created two duotone 500x500 images and blurred them to get soft transitions between regions.
The numbers show the sum of absolute differences between the estimated alpha mattes and the ground truth.
Closed-form matting~\cite{closedformpami} uses local information flow, KNN Matting~\cite{knnpami} uses HSV- or RGB-based similarity measure, and manifold-preserving edit propagation~\cite{manifold} uses LLE weights~\cite{lle}.
We observe a performance improvement in large opacity gradients even when only
the color-mixture flow (CMF) is used (Section~\ref{sec:mixture}). Notice also that both large gradients and holes are recovered with high performance using our final formulation.
See text for further discussion.
\negvspace
}
\label{fig:simpleExamplesColor}
\end{figure*}

The main application of natural image matting is compositing, \ie combining different scenes together to generate a new image.
Image matting methods aim to provide accurate opacities such that when the foreground is overlayed onto a novel background, the transitions between them look natural.
However, together with the matte, compositing requires the actual, \emph{unmixed} layer colors for realistic composites.
The layer colors appear as a mixture of foreground and background colors in the input image, and they are underconstrained even with a given matte.
Hence, accurate estimation of the layer colors is a critical component of a compositing pipeline and still an active research problem.

\emph{Affinity-based} methods \cite{closedformpami,knnpami,manifold} constitute one of the prominent natural matting approaches in literature.
These methods make use of pixel similarities to propagate the alpha values from the known-alpha regions to the unknown region.
They provide a clear mathematical formulation, can be solved in closed-form, are easy to implement, and typically produce spatially consistent mattes.
In addition, due to their formulation that can be modeled as a graph structure with each pixel as a node, affinity-based approaches can be generalized to related applications such as layer color estimation~\cite{closedformpami}, edit propagation~\cite{manifold}, and soft segmentation~\cite{spectralpami}.
Studying affinity-based approaches for natural matting can open new directions for a larger set of applications in the image processing community.

In spite of these advantages, current affinity-based methods fail to effectively handle alpha gradients spanning large areas and spatially disconnected regions (i.e. \emph{holes}) even in simple cases as demonstrated in Figure~\ref{fig:simpleExamplesColor}.
This is because a straightforward formulation using the pixel-to-pixel affinity definitions can not effectively represent the complex structures that are commonly seen in real-life objects.
In order to alleviate these shortcomings, we rely on a careful, case-by-case design of how alpha values should propagate inside the image.
We conceptualize the affinities as \emph{information flows} to help understanding and designing effective graph-based structures to propagate information in the image.
We define several information flows, some of which target unknown-opacity regions that are remote and hence does not receive enough information in previous formulations. Other types of information flows address issues such as evenly distributing information inside the unknown region.
We formulate this strategy through the use of a variety of affinity definitions including the \emph{color-mixture flow}, which is based on local linear embedding and tailored for image matting.
Step-by-step improvements on the matte quality as we gradually add new building blocks of our information flow strategy are illustrated in Figure~\ref{fig:teaser}.
Our final linear system can be solved in closed-form and results in a significant quality improvement over the state-of-the-art.
We demonstrate the matting quality improvement quantitatively, as well as through a visual inspection of challenging image regions.
We also show that our energy function can be reformulated as a post-processing step for regularizing the spatially inconsistent mattes estimated by sampling-based natural matting algorithms.

This document is an extended version of our CVPR publication~\cite{ifm}.
In this extended version, we additionally
(i) propose a novel foreground color estimation formulation where we introduce a new form of local information flow,
(ii) demonstrate that our method achieves state-of-the-art quality in green-screen keying,
(iii) provide an in-depth spectral analysis of individual forms of information flow,
and (iv) present a discussion on how our method relates to sampling-based matting methods, 
as well as new quantitative and qualitative results.

\section{Related work}
\label{sec:related}


\noindent\textbf{Natural Image Matting \quad}
The numerous natural matting methods in the literature can be mainly categorized as sampling-based, learning-based or affinity-based.
We briefly review the most relevant here and refer the reader to a comprehensive survey~\cite{Zhu2015} for further information.

Sampling-based methods \cite{csc,kldiv,comprehensive,shared} typically seek to gather numerous samples from the background and foreground regions defined by the trimap and select the best-fitting pair according to their individually defined criteria for representing an unknown pixel as a mixture of foreground and background.
While they perform well especially around remote and challenging structures, they require affinity-based regularization to produce spatially consistent mattes.
Also, their methodology typically focuses solely on matting and they typically can not generalize any other applications unlike the affinity-based counterparts.

Machine learning has been used to aid in estimating the alpha in a semi-supervised setting~\cite{learningbased}, to estimate a trimap in constrained settings~\cite{deepportrait} or to combine results of other matting methods for a better matte~\cite{dcnn}.
Recently, a deep neural network architecture has been proposed~\cite{deepmatting} that generates high-quality mattes with the help of semantic knowledge that can be extracted from the image.
In order to train such a network, Xu~\etal\cite{deepmatting} generated a dataset of 50k images with ground-truth mattes.
Our method outperforms all current learning-based methods in the alpha matting benchmark~\cite{alphabenchmark} despite not taking advantage of a large dataset with labels. We hope that our formulation and the concepts presented in the paper will inspire next-generation learning-based matting methods. 

Affinity-based matting methods mainly make use of pixel similarity metrics that rely on color similarity or spatial proximity and propagate the alpha values from regions with known opacity.
Local affinity definitions, prominently the matting affinity~\cite{closedformpami}, operate on a local patch around the pixel to determine the amount of information flow and propagate alpha values accordingly.
The matting affinity is also adopted in a post-processing step in most sampling-based methods as proposed by Gastal and Oliveira~\cite{shared}.

Methods utilizing nonlocal affinities similarly use color similarity and spatial proximity for determining how the alpha values of different pixels should relate to each other.
KNN matting~\cite{knnpami} determines several neighbors for every unknown pixel and enforces them to have similar alpha values relative to their distance in a feature space.
The manifold-preserving edit propagation algorithm~\cite{manifold} also determines a set of neighbors for every pixel but represents each pixel as a linear combination of its neighbors in their feature space.

Chen~\etal\cite{lnsp} proposed a hybrid approach that uses the sampling-based robust matting~\cite{robust} as a starting point and refines its outcome through a graph-based technique where they combine a nonlocal affinity~\cite{manifold} and the matting affinity.
Cho \etal\cite{dcnn} combined the results of closed-form matting~\cite{closedformpami} and KNN matting~\cite{knnpami}, as well as the sampling-based method comprehensive sampling~\cite{comprehensive}, by feeding them into a convolutional neural network.

In this work, we propose color-mixture flow and discuss its advantages over the affinity definition utilized by Chen~\etal\cite{manifold}.
We also define three other forms of information flow, which we use to carefully distribute the alpha information inside the unknown region.
Our approach differs from Chen~\etal\cite{lnsp} in that our information flow strategy goes beyond combining various pixel affinities, as we discuss further in Section~\ref{sec:method}, while requiring much less memory to solve the final system.
Instead of using the results of other affinity-based methods directly as done by Cho~\etal\cite{dcnn}, we formulate an elegant formulation that has a closed-form solution.
To summarize, we present a novel, purely affinity-based matting algorithm that generates high-quality alpha mattes without making use 
of sampling or a learning step.

\noindent\textbf{Layer Color Estimation \quad}
For a given alpha matte, the corresponding foreground colors should also be estimated before compositing.
Although the alpha matte is assumed to be given for the foreground color estimation, the problem is still underconstrained as there are 6 unknowns and 3 equations.
Levin~\etal\cite{closedformpami} use the gradient of the alpha matte as a spatial smoothness measure and formulate the layer color estimation as a linear problem.
Using only a smoothness measure limits their performance especially in remote regions of the foreground.
Chen~\etal\cite{knnpami} use the color-similarity measure they employ for matte estimation also for layer color estimation.
Typically, using only a color-similarity metric results in incorrectly flat-colored regions and suppressed highlight colors in the foreground.
In this work, we introduce a second spatial smoothness measure for the layer colors.
We use in total 4 forms of information flow together for the layer estimation and show that our linear system improves the layer color quality especially in remote parts of the matte.

\noindent\textbf{Green-Screen Keying \quad}
A more constrained version of the natural image matting problem is referred as green-screen keying, where the background colors are homogeneous in a controlled setting.
While this problem can be seen as a simpler version of natural matting, as green-screen keying is heavily utilized in professional production~\cite{keying}, the expected quality of the results is immense.
In the movie post-production industry, multiple commercial software such as Keylight or Primatte are used by professional graphical artists to get high-quality keying results.
These software typically use chroma-based or luma-based algorithms and provide many parameters that help the artist tweak the results.
In their early work, Smith and Blinn~\cite{Smith96} formulate the use of the compositing equation for a fixed background color. 
Recently, an unmixing-based green-screen keying method has been proposed~\cite{keying} that uses a global color model of the scene and a per-pixel nonlinear energy function to extract the background color in high precision.
In their paper, they compare their method to state-of-the-art natural matting methods and show that the current matting methods fail to give acceptable results in green-screen settings.
In this paper, we show that our matting and color estimation methods outperform the natural matting methods and generate comparable results to that of specialized keying methods or commercial software without any parameter tweaking. 

\section{Method}
\label{sec:method}

\newcommand{\fgt}{$\mathcal{F}$}
\newcommand{\bgt}{$\mathcal{B}$}
\newcommand{\unt}{$\mathcal{U}$}
\newcommand{\knt}{$\mathcal{K}$}
\newcommand{\wholet}{$\mathcal{I}$}

\newcommand{\fgm}{\mathcal{F}}
\newcommand{\bgm}{\mathcal{B}}
\newcommand{\unm}{\mathcal{U}}
\newcommand{\knm}{\mathcal{K}}
\newcommand{\wholem}{\mathcal{I}}

Trimaps are typically given as user input in natural matting, and they consist of three regions: fully opaque (foreground), fully transparent (background) and of unknown opacity.
\fgt, \bgt\, and \unt\ will respectively denote these regions, and
\knt\ will represent the union of \fgt\ and \bgt.
Affinity-based methods operate by propagating opacity information from \knt\ into \unt\ using a variety of affinity definitions.
We define this flow of information in multiple ways so that each pixel in \unt\ receives information effectively from different regions in the image.

The opacity transitions in a matte occur as a result of the original colors in
the image getting mixed with each other due to transparency or intricate parts of an object.
We make use of this fact by representing each pixel in \unt\ as a mixture of
similarly-colored pixels and defining a form of information flow that we call
\emph{color-mixture flow} (Section~\ref{sec:mixture}).
We also add connections from every pixel in \unt\ to both \fgt\ and \bgt\ to facilitate direct information flow from known-opacity regions to even the most remote opacity-transition regions in the image (Section~\ref{sec:knownToUnknown}).
In order to distribute the information from the color-mixture and \knt-to-\unt\ flows, we define intra-\unt\ flow of information, where pixels with similar colors inside \unt\ share information on their opacity with each other (Section~\ref{sec:insideUnknown}).
Finally, we add local information flow, a pixel affecting the opacity of its
immediate spatial neighbors, which ensures spatially coherent end results (Section~\ref{sec:local}).
We formulate the individual forms of information flow as energy functions and aggregate them in a global optimization formulation (Section~\ref{sec:linearSystem}).


\begin{figure*}
{
\footnotesize
\begin{tabular}
{   K{0.175\linewidth}
    K{0.175\linewidth}
    K{0.175\linewidth}
    K{0.175\linewidth}
    K{0.175\linewidth}
}
Input&
Ground-truth&
Without \knt-to-\unt\ flow&
Without confidences ($\eta_p$)&
Our method
\end{tabular}
}
\showimagew[\linewidth]{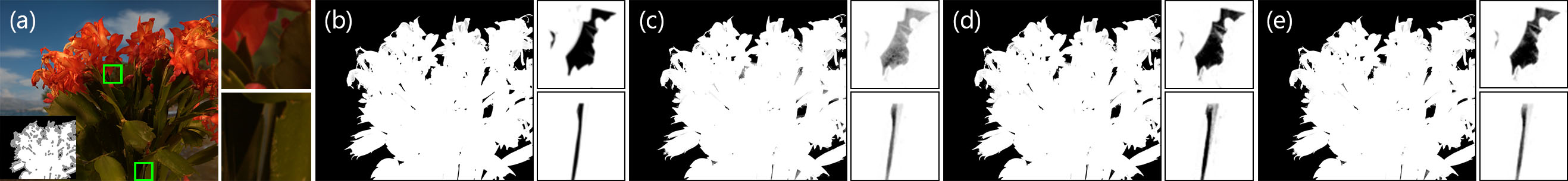}
\caption{
Direct information flow from both \fgt\ and \bgt\ to even the most remote regions in \unt\ increases our performance around holes significantly (top inset).
Using confidences further increases the performance, especially around regions where FG and BG colors are similar (bottom inset).
}
\negvspace
\label{fig:Training21Steps}
\end{figure*}

\subsection{Color-mixture information flow}
\label{sec:mixture}

Due to transparent objects as well as fine structures and sharp edges of an object that cannot be fully captured due to the finite-resolution of the imaging sensors, certain pixels of an image inevitably contain a mixture of corresponding foreground and background colors.
By investigating these color mixtures, we can derive an important clue on how to propagate alpha values between pixels.
The amount of the original foreground color in a particular mixture determines the opacity of the pixel.
Following this fact, if we represent the color of a pixel as a weighted
combination of the colors of several others, those weights should correspond to the opacity relation between the pixels.

In order to make use of this relation, for every pixel in \unt, we find $K_{CM}=20$ similar pixels in a feature space by an approximate K nearest neighbors search in the whole image.
We define the feature vector for this search as $[r, g, b, \tilde{x},
\tilde{y}]^T$, where $\tilde{x}$ and $\tilde{y}$ are the image coordinates
normalized by image width and height, and the rest are the RGB values of the pixel.
This set of neighbors, selected as similar-colored pixels that are also close-by, is denoted by $\mathcal{N}_p^{CM}$.

We then find the weights of the combination $w^{CM}_{p,q}$ that will determine the amount of information flow between the pixels $p$ and  $q \in \mathcal{N}_p^{CM}$.
The weight of each neighbor is defined such that the weighted combination of their colors yields the color of the original pixel:
\begin{equation}
\argmin_{w^{CM}_{p,q}}\left\| \ve{c}_p - \sum_{q\in\mathcal{N}_p^{CM}} w^{CM}_{p,q} \ve{c}_q \right\| ^ 2,
\label{eq:lle}
\end{equation}
where $\ve{c}_p$ represents the 3x1 vector of RGB values.
We minimize this energy using the method by Roweis and Saul~\cite{lle}.
Note that since we are only using RGB values, the neighborhood correlation matrix computed during the minimization has a high chance of being singular as there could easily be two neighbors with identical colors.
So, we condition the neighborhood correlation matrix by adding $10^{-3}I_{K_{CM}\times K_{CM}}$ to it before inversion, where $I_{K_{CM}\times K_{CM}}$ is the identity matrix.

Note that while we use the method by Roweis and Saul~\cite{lle} to minimize the energy in \refeq{eq:lle}, we do not fully adopt their local linear embedding (LLE) method.
LLE finds a set of neighbors in a feature space and uses all the variables in the feature space to compute the weights in order to reduce the dimentionality of input data.
Manifold-preserving edit propagation~\cite{manifold} and LNSP matting~\cite{lnsp} algorithms make use of the LLE weights directly in their formulation for image matting.
However, since we are only interested in the weighted combination of colors and not the spatial coordinates, we exclude the spatial coordinates in the energy minimization step.
This increases the validity of the estimated weights, effects of which can be observed even in the simplest cases such as in Figure~\ref{fig:simpleExamplesColor}, where manifold-preserving weight propagation and CMF-only results only differ in the weight computation step.

The energy term for the color-mixture flow is defined as:
\begin{equation}
E_{CM} = \sum_{p \in \unm} \left( \alpha_p - \sum_{q \in \mathcal{N}_p^{CM}} w^{CM}_{p,q} \alpha_q \right)^2.
\label{eq:colormixtureenergy}
\end{equation}
%


\subsection{\knt-to-\unt\ information flow}
\label{sec:knownToUnknown} 

The color-mixture flow already provides useful information on how the mixed-color pixels are formed.
However, many pixels in \unt\ receive information present in the trimap indirectly through their neighbors, all of which can possibly be in \unt.
This indirect information flow might not be enough especially for remote regions that are far away from \knt.

In order to facilitate the flow of information from both \fgt\ and \bgt\ directly into every region in \unt, we add connections from every pixel in \unt\ to several pixels in \knt.
For each pixel in \unt, we find $K_{\knm\unm} = 7$ similar pixels in both \fgt\ and \bgt\ separately to form the sets of pixels $\mathcal{N}_p^\fgm$ and $\mathcal{N}_p^\bgm$ with K nearest neighbors search using the feature space $[r, g, b, 10*\tilde{x}, 10*\tilde{y}]^T$ to favor close-by pixels.
We use the pixels in $\mathcal{N}_p^\fgm$ and $\mathcal{N}_p^\bgm$ together to represent the pixel color $\ve{c}_p$ by minimizing the energy in \refeq{eq:lle}.
Using the resulting weights $w^\fgm_{p,q}$ and $w^\bgm_{p,q}$, we define an energy function to represent the \knt-to-\unt\ flow:
\begin{equation}
E_{\knm\unm} = \sum_{p \in \unm} \left( \alpha_p - \sum_{q \in \mathcal{N}_p^\fgm} w^\fgm_{p,q} \alpha_q - \sum_{q \in \mathcal{N}_p^\bgm} w^\bgm_{p,q} \alpha_q \right)^2
\label{eq:knowntounknownenergy}
\end{equation}
Note that $\alpha_q = 1$ for $q \in \fgm$ and $\alpha_q = 0$ for $q \in \bgm$.
This fact allows us to define two combined weights, one connecting a pixel to \fgt\ and another to \bgt, as:
\begin{equation}
w^\fgm_p = \sum_{q \in \mathcal{N}_p^\fgm} w^\fgm_{p,q}
\quad \text{and} \quad
w^\bgm_p = \sum_{q \in \mathcal{N}_p^\bgm} w^\bgm_{p,q}
\label{eq:fgweights}
\end{equation}
such that $w^\fgm_p + w^\bgm_p = 1$, and rewrite \refeq{eq:knowntounknownenergy} as:
\begin{equation}
E_{\knm\unm} = \sum_{p \in \unm} \left( \alpha_p - w^\fgm_p \right)^2.
\label{eq:knowntounknownenergy2}
\end{equation}

The energy minimization in \refeq{eq:lle} gives us similar weights for all $q$ when $\ve{c}_q$ are similar to each other.
As a result, if $\mathcal{N}^\fgm_p$ and $\mathcal{N}^\bgm_p$ have pixels with similar colors, the estimated weights $w^\fgm_p$ and $w^\bgm_p$ become unreliable.
We account for this fact by augmenting the energy function in \refeq{eq:knowntounknownenergy2} with confidence values.

We can determine the colors contributing to the mixture estimated by \refeq{eq:lle} using the weights $w^\fgm_{p,q}$ and $w^\bgm_{p,q}$:
\begin{equation}
\ve{c}_{p}^{\fgm} = \frac
{\sum_{q \in \mathcal{N}_p^{\fgm}} w^\fgm_{p,q} \ve{c}_q}
{w^\fgm_p},
\quad
\ve{c}_{p}^{\bgm} = \frac
{\sum_{q \in \mathcal{N}_p^\bgm} w^\bgm_{p,q} \ve{c}_q}
{w^\bgm_p},
\label{eq:fgbgcolors}
\end{equation}
and define a confidence metric according to how similar the estimated foreground color $\ve{c}_{p}^{\fgm}$ and background color $\ve{c}_{p}^{\bgm}$ are:
\begin{equation}
\eta_p = \left\| \ve{c}_{p}^{\fgm} - \ve{c}_{p}^{\bgm}\right\|^2 / 3.
\label{eq:confidence}
\end{equation}
The division by 3 is to get the confidence values between $[0,1]$.
We update the new energy term to reflect our confidence in the estimation:
\begin{equation}
\tilde{E}_{\knm\unm} = \sum_{p \in \unm} \eta_p \left( \alpha_p - w^\fgm_p\right)^2.
\label{eq:combinationEnergy4}
\end{equation}
This update to the energy term increases the matting quality in regions with similar foreground and background colors, as seen in \reffig{fig:Training21Steps}.

\begin{figure}
{
\footnotesize
\begin{tabular}
{   K{0.285\linewidth}
    K{0.285\linewidth}
    K{0.285\linewidth}
}
Input&
No \knt-to-\unt\ flow&
With \knt-to-\unt\ flow
\end{tabular}
}
\showimagew[\linewidth]{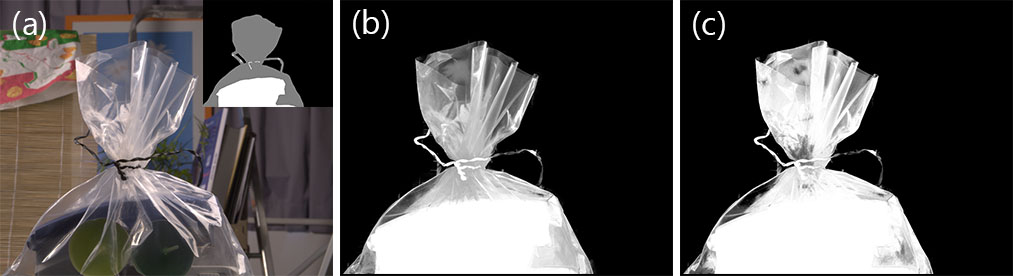}
\caption{
\knt-to-\unt\ flow does not perform well when the foreground object is highly-transparent.
See text for discussion.
}
\negvspace
\label{fig:E1E2}
\end{figure}

It should be noted that the \knt-to-\unt\ flow is not reliable when the foreground is highly transparent, as seen in \reffig{fig:E1E2}.
This is mainly due to the low representational power of $\mathcal{N}^\fgm_p$ and
$\mathcal{N}^\bgm_p$ for $\ve{c}_p$ around large highly-transparent regions as
the nearest neighbors search does not give us well-fitting pixels for
$w^\fgm_{p,q}$ estimation.
We construct our final linear system accordingly in Section~\ref{sec:linearSystem}.


\subsubsection{Pre-processing the trimap}

Prior to determining $\mathcal{N}^\fgm_p$ and $\mathcal{N}^\bgm_p$, we pre-process the input trimap in order to facilitate finding more reliable neighbors, which in turn increases the effectiveness of the \knt-to-\unt\ flow.
Trimaps usually have regions marked as \unt\ despite being fully opaque or
transparent, as drawing a very detailed trimap is both cumbersome and prone to errors.

\begin{figure}{
\footnotesize
\begin{tabular}
{   K{0.15\linewidth}
    K{0.15\linewidth}
    K{0.15\linewidth}
    K{0.15\linewidth}
    K{0.15\linewidth}
}
Input&
Trimap&
No trim&
CS trim&
Both trims
\end{tabular}
}
\showimagew[\linewidth]{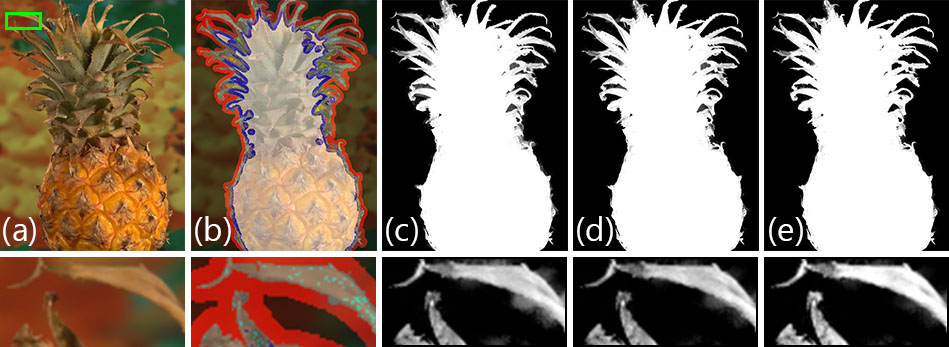}
\caption{
The trimap is shown overlayed on the original image (b) where the extended foreground regions are shown with blue (CS trimming~\cite{comprehensive}) and cyan (patch-search) and the extended background regions with red (CS trimming) and yellow (patch-search).
CS trimming makes the fully opaque / transparent regions cleaner, while our trimming improves the results around remote structures.
}
\negvspace
\negvspace
\label{fig:Test5Steps}
\end{figure}

Several methods \cite{csc,kldiv} refine the trimap as a pre-processing step by expanding \fgt\ and \bgt\ starting from their boundaries with \unt\ as proposed by Shahrian~\etal\cite{comprehensive}.
Incorporating this technique improves our results as shown in \reffig{fig:Test5Steps}(d).
We also apply this extended \fgt\ and \bgt\ regions after the matte estimation as a post-processing.
Since this trimap trimming method propagates known regions only to nearby pixels, in addition to this edge-based trimming, we also make use of a patch-based trimming step.

To this end, we extend the transparent and opaque regions by relying on patch statistics.
We fit a 3D RGB normal distribution $N_p$ to the $3\times3$ window around each pixel $p$.
In order to determine the most similar distribution in \fgt\ for a pixel $p \in \unm$, we first find the 20 distributions with closest mean vectors.
We define the foreground match score $b^{\fgm}_p = \min_{q \in \fgm} B(N_p, N_q)$, where $B(\cdot, \cdot)$ represents the Bhattacharyya distance between two distributions.
We find the match score for background $b^{\bgm}_p$ the same way.
We then select a region for pixel $p$ according to the following rule:
\begin{equation}
p \in 
\begin{cases}
\hat{\mathcal{F}} \quad \text{if\ } b^{\fgm}_p < \tau_c \text{\ and\ } b^{\bgm}_p > \tau_f
\\
\hat{\mathcal{B}} \quad \text{if\ } b^{\bgm}_p < \tau_c \text{\ and\ } b^{\fgm}_p > \tau_f
\\
\hat{\mathcal{U}} \quad \text{otherwise}
\end{cases}
\label{eq:trimdecide}
\end{equation}
Simply put, an unknown pixel is marked as $\hat{\mathcal{F}}$, \ie in foreground after trimming, if it has a strong match in \fgt\ and no match in \bgt, which is determined by constants $\tau_c = 0.25$ and $\tau_f = 0.9$.
By inserting known-alpha pixels in regions far away from \unt-\knt\ boundaries, we further increase the matting performance in challenging remote regions (\reffig{fig:Test5Steps}(e)).

\begin{figure*}
{
\footnotesize
\begin{tabular}
{   K{0.175\linewidth}
    K{0.175\linewidth}
    K{0.175\linewidth}
    K{0.175\linewidth}
    K{0.175\linewidth}
}
Input&
Ground-truth&
Sampling-based $\hat\alpha$ \cite{comprehensive}&
Regularization by \cite{shared}&
Our regularization
\end{tabular}
}
\showimagew[\linewidth]{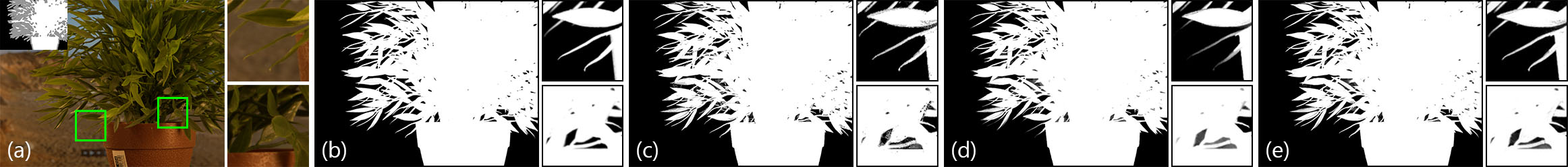}
\caption{
The matte regularization method by Gastal and Oliveira~\cite{shared} loses remote details (top inset) or fills in holes (bottom inset) while our regularization method is able to preserve these details caught by the sampling-based method.
}
\negvspace
\negvspace
\label{fig:MatteRefinement}
\end{figure*}


\subsection{Intra-\unt\ information flow}
\label{sec:insideUnknown}

Each individual pixel in \unt\ receives information through the color-mixture and \knt-to-\unt\ flows.
In addition to these, we would like to distribute the information inside \unt\ effectively.
We achieve this by encouraging pixels with similar colors inside \unt\ to have similar opacity.

For each pixel in \unt, we find $K_\unm = 5$ nearest neighbors only inside \unt\ to determine $\mathcal{\hat{N}}_p^\unm$ using the feature vector defined as $\ve{v} = [r, g, b, \tilde{x} / 20, \tilde{y} / 20]^T$.
Notice that we scale the coordinate members of the feature vector we used in Section~\ref{sec:mixture} to decrease their effect on the nearest neighbor selection.
This lets $\mathcal{\hat{N}}_p^\unm$ have pixels inside \unt\ that are far away, so that the information moves more freely inside the unknown region.
We use the neighborhood \mbox{$\mathcal{N}_p^\unm = \mathcal{\hat{N}}_p^\unm \cup \{q\ |\ p \in  \mathcal{\hat{N}}_q^\unm\}$} to make sure that information flows both ways between $p$ to $q\in{\mathcal{\hat{N}}_p^\unm}$.
We then determine the amount of information flow using the $L^1$ distance between feature vectors:
\begin{equation}
w_{p,q}^\unm = \max\left(1 - \left\| \ve{v}_p - \ve{v}_q \right\|_1,\ 0\right) \quad \forall q \in \mathcal{N}_p^\unm.
\label{eq:knnweight}
\end{equation}
The energy term for intra-\unt\ flow then can be defined as:
\begin{equation}
E_{\unm\unm} = \sum_{p \in \unm} \sum_{q \in \mathcal{N}_p^\unm} w^\unm_{p,q} \left( \alpha_p - \alpha_q \right)^2.
\label{eq:similarityEnergy}
\end{equation}
The information sharing between the unknown pixels increases the matte quality
around intricate structures as demonstrated in Figure~\ref{fig:teaser}(d).

KNN matting~\cite{knnpami} uses a similar affinity definition to make similar-color pixels have similar opacities.
However, relying only on this form of information flow for the whole image creates some typical artifacts in the matte.
Depending on the feature vector definition and the image colors, the matte may erroneously underrepresent the smooth transitions (KNN - HSV case in \reffig{fig:simpleExamplesColor}) when the neighbors of the pixels in \unt\ happen to be mostly in only \fgt\ or \bgt, or create flat alpha regions instead of subtle gradients (KNN - RGB case in \reffig{fig:simpleExamplesColor}).
Restricting information flow to be solely based on color similarity fails to
represent the complex alpha transitions or wide regions with an alpha gradient.


\subsection{Local information flow}
\label{sec:local}

Spatial connectivity is one of the main cues for information flow.
We connect each pixel in \unt\ to its 8 immediate neighbors denoted by $\mathcal{N}_p^L$ to ensure spatially smooth mattes.
The amount of local information flow should also adapt to strong edges in the image.

To determine the amount of local flow, we rely on the matting affinity definition proposed by Levin~\etal\cite{closedformpami}.
The matting affinity utilizes the local patch statistics to determine the weights $w^L_{p,q}$, $q \in \mathcal{N}_p^L$.
We define our related energy term as follows:
\begin{equation}
E_L = \sum_{p \in \unm} \sum_{q \in \mathcal{N}_p^L} w^L_{p,q} \left( \alpha_p - \alpha_q \right)^2.
\label{eq:localEnergy}
\end{equation}
Despite representing local information flow well, matting affinity by itself fails to represent large transition regions (\reffig{fig:simpleExamplesColor} top), or isolated regions that have weak or no spatial connection to \fgt\ or \bgt\ (\reffig{fig:simpleExamplesColor} bottom).


\subsection{Linear system and energy minimization}
\label{sec:linearSystem}

Our final energy function is a combination of the four energies representing the individual information flows:
\begin{equation}
E_1 = E_{CM} + \sigma_{\knm\unm} E_{\knm\unm} + \sigma_{\unm\unm} E_{\unm\unm} + \sigma_L E_L + \lambda E_\mathcal{T},
\label{eq:energy}
\end{equation}
where $\sigma_{\knm\unm} = 0.05$, $\sigma_{\unm\unm} = 0.01$, $\sigma_{L} = 1$ and $\lambda = 100$ are algorithmic constants determining the strength of corresponding information flows, and
\begin{equation*}
E_\mathcal{T} = \sum_{p \in \fgm} \left(\alpha_p - 1\right) ^ 2 + \sum_{p \in \bgm} (\alpha_p - 0) ^ 2
\end{equation*}
is the energy term to keep the known opacity values constant.
For an image with $N$ pixels, by defining $N \times N$ sparse matrices $W_{CM}$, $W_{\unm\unm}$ and $W_{L}$ that have non-zero elements for the pixel pairs with corresponding information flows and the vector $\ve{w}^\fgm$ that has elements $w^\fgm_p$ for $p\in\unm$, $1$ for $p\in\fgm$ and $0$ for $p\in\bgm$, we can write \refeq{eq:energy} in matrix form as:
\begin{equation}
\begin{split}
E_1 =        
                &            \ve\alpha^T  \mathcal{L}_{IFM} \ve\alpha + 
                              (\ve\alpha - \ve{w}^\fgm)^T \sigma_{\knm\unm} \mathcal{H} (\ve\alpha - \ve{w}^\fgm) + \\
                &             (\ve\alpha - \ve\alpha_\knm)^T \lambda \mathcal{T} (\ve\alpha - \ve\alpha_\knm),
\label{eq:energy1matrix}
\end{split}
\end{equation}
where $\mathcal{T}$ is an $N \times N$ diagonal matrix with diagonal entry ($p$, $p$) 1 if $p \in \knm$ and 0 otherwise, 
$\mathcal{H}$ is a sparse matrix with diagonal entries $\eta_p$ as defined in \refeq{eq:confidence},
$\ve\alpha_\knm$ is a row vector with $p^{\text{th}}$ entry being 1 if $p \in \fgm$ and 0 otherwise, 
$\ve\alpha$ is a row-vector of the alpha values to be estimated, and
$\mathcal{L}_{IFM}$ is defined as:
\begin{equation}
\begin{split}
\mathcal{L}_{IFM} 
=
&
(D_{CM} - W_{CM})^T (D_{CM} - W_{CM}) + 
\\
& 
\sigma_{\unm\unm} (D_{\unm\unm} - W_{\unm\unm}) + 
\sigma_L (D_L - W_L),
\end{split}
\end{equation}
where the diagonal matrix $D_{(\cdot)}(i,i) = \sum_jW_{(\cdot)}(i,j)$.

The energy in \refeq{eq:energy1matrix} can be minimized by solving
\begin{equation}
\left( \mathcal{L}_{IFM} + \lambda \mathcal{T} + \sigma_{\knm\unm} \mathcal{H} \right) \ve\alpha =
 \left(\lambda \mathcal{T} + \sigma_{\knm\unm} \mathcal{H} \right) \ve{w}^\fgm.
\label{eq:energy1solution}
\end{equation}

We define a second energy function that excludes the \knt-to-\unt\ information flow:
\begin{equation}
E_2 = E_{CM} + \sigma_{\unm\unm} E_{\unm\unm} + \sigma_L E_L + \lambda E_\mathcal{T},
\label{eq:energy2}
\end{equation}
which can be written in matrix form as:
\begin{equation}
E_2 =        
\ve\alpha^T  \mathcal{L}_{IFM} \ve\alpha + 
(\ve\alpha - \ve\alpha_\knm)^T \lambda \mathcal{T} (\ve\alpha - \ve\alpha_\knm),
\label{eq:energy2matrix}
\end{equation}
and can be minimized by solving:
\begin{equation}
\left( \mathcal{L}_{IFM} + \lambda \mathcal{T} \right) \ve\alpha = \lambda \mathcal{T} \ve\alpha_\knm.
\label{eq:energy2solution}
\end{equation}

We solve the linear systems of equations in \refeq{eq:energy1solution} and \refeq{eq:energy2solution} using the preconditioned conjugate gradients method~\cite{pcg}.


As mentioned before, the \knt-to-\unt\ information flow is not effective for highly transparent objects.
To determine whether to include the \knt-to-\unt\ information flow and solve for $E_1$, or to exclude it and solve for $E_2$ for a given image, we use a simple histogram-based classifier to determine if we expect a highly transparent result.

If the matte is highly transparent, the pixels in \unt\ are expected to mostly have colors that are a mixture of \fgt\ and \bgt\ colors.
On the other hand, if the true alpha values are mostly 0 or 1 except for soft transitions, the histogram of \unt\ will likely be a linear combination of the histograms of \fgt\ and \bgt\ as \unt\ will mostly include very similar colors to that of \knt.
Following this observation, we attempt to express the histogram of the pixels in \unt, $\mathcal{D}_\unm$, as a linear combination of $\mathcal{D}_\fgm$ and $\mathcal{D}_\bgm$.
The histograms are computed from the 20 pixel-wide region around \unt\ in \fgt\ and \bgt, respectively.
We define the error $e$, the metric of how well the linear combination represents the true histogram, as:
\begin{equation}
e = \min_{a,b}\| a \mathcal{D}_\fgm + b \mathcal{D}_\bgm - \mathcal{D}_\unm \| ^2.
\label{eq:histLinComb}
\end{equation}
%
Higher $e$ values indicate a highly-transparent matte, in which case we prefer $E_2$ over $E_1$.

\begin{figure*}
{
\footnotesize
\begin{tabular}
{   K{0.120\linewidth}
    K{0.150\linewidth}
    K{0.150\linewidth}
    K{0.150\linewidth}
    K{0.150\linewidth}
    K{0.130\linewidth}
}
Input image&
Ground truth&
Only $\alpha$-transition&
Both local flows&
Color-mix. \& local&
Our result
\end{tabular}
}
\showimagew[\linewidth]{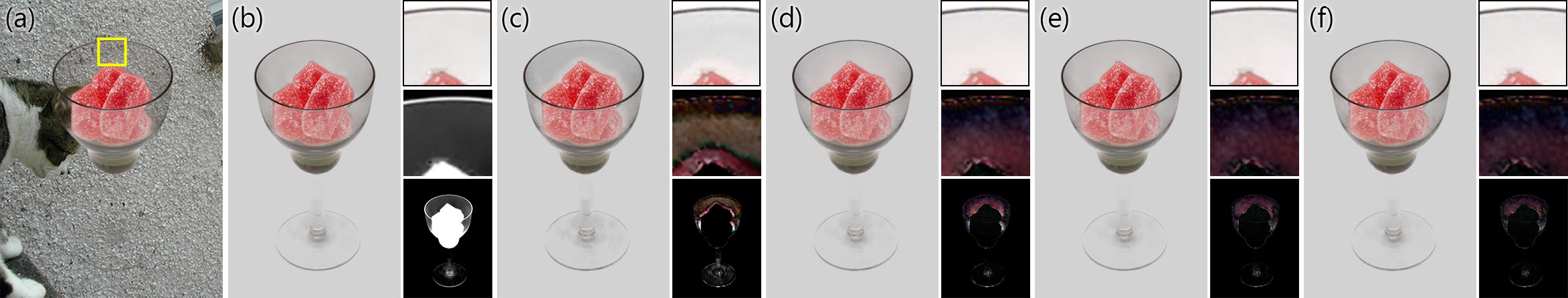}
\caption{
Color estimation results using a growing set of information flows using the ground truth matte.
The bottom-right in each set shows per-pixel absolute difference between the estimation and ground truth multiplied by ten.
See text for discussion.
}
\negvspace
\negvspace
\label{fig:stepByStep_color}
\end{figure*}

\section{Matte regularization for sampling-based matting methods}
\label{sec:postprocessing}

Sampling-based natural matting methods usually select samples for each pixel in
\unt\ either independently or by paying little attention to spatial coherency.
In order to obtain a spatially coherent matte, the common practice is to combine their initial guesses for alpha values with a smoothness measure.
Multiple methods \cite{csc,shared,kldiv,comprehensive} adopt the post-processing method proposed by Gastal and Oliveira~\cite{shared} which combines the matting affinity \cite{closedformpami} with the sampling-based alpha values and corresponding confidences.
This post-processing technique leads to improved mattes, but since it involves only local smoothness, the results can still be suboptimal as seen in \reffig{fig:MatteRefinement}(d).

Our approach with multiple forms of information flow can also be used for post-processing in a way similar to that of Gastal and Oliveira~\cite{shared}.
Given the initial alpha values $\hat\alpha_p$ and confidences $\hat\eta_p$ found by a sampling-based method, we define the matte regularization energy:
\begin{equation}
E_R = E_2 + \sigma_R \sum_{p \in \unm} \hat\eta_p ( \alpha_p - \hat\alpha_p)^2,
\label{eq:regularization}
\end{equation}
where $\sigma_R = 0.05$ determines how much loyalty should be given to the initial values.
This energy can be written in matrix form and solved as a linear system in the same way we did in Section~\ref{sec:linearSystem}.

%
\reffig{fig:MatteRefinement} shows that this non-local regularization of mattes is more effective especially around challenging foreground structures such as long leaves or holes as seen in the insets.
In Section~\ref{sec:results:regularization}, we will numerically explore the improvement we achieve by replacing the matte regularization step with ours in several sampling-based methods.

\section{Foreground color estimation}
\label{sec:colorest}

In addition to the alpha matte, we need the \emph{unmixed} foreground colors~\cite{keying} that got into the color mixture in transition pixels for seamlessly compositing the foreground onto a novel background. 
Similar to Levin~\etal\cite{closedformpami} and Chen~\etal\cite{knnpami}, we estimate the foreground colors for a given matte, after the matte estimation.

We propagate the layer colors from opaque and transparent regions in a similar way we propagate known alpha values in Section~\ref{sec:method}.
We make use of the color-mixture and the intra-\unt\ information flows by extending the search space and affinity computation to include the given alpha values together with spatial coordinates and pixel colors.
We also use the spatial smoothness measure proposed by Levin~\etal\cite{closedformpami} in addition to a second spatial smoothness measure we introduce in this paper.
Figure~\ref{fig:stepByStep_color} shows how our color estimation result improves as we add more forms of information flow.

\subsection{Information flow definitions}

In the layer color estimation problem, the input is assumed to be the original image together with an alpha matte.
This requires us to redefine the three regions using the matte instead of a trimap:
\begin{equation}
p \in 
\begin{cases}
\tilde{\fgm} \quad \text{if\ } \tilde\alpha_p = 1 \\
\tilde{\bgm} \quad \text{if\ } \tilde\alpha_p = 0 \\
\tilde{\unm} \quad \text{otherwise.}
\end{cases}
\end{equation}
$\tilde\alpha_p$ denote the alpha values that are given as input.
The foreground and background colors to be estimated will be denoted by $\ve{f}$ and $\ve{b}$.
For a pixel $p$, the compositing equation we would like to satisfy can be written as:
\begin{equation}
\ve{c}_p = \tilde\alpha_p \ve{f}_p + \left( 1 -\tilde \alpha_p \right) \ve{b}_p
\label{eq:compositing}
\end{equation}
We will formulate the energy functions for a single color channel and solve for red, green and blue channels independently.
The scalars $f$ and $b$ will denote the values for a single color channel.


\subsubsection{Local information flows}
\label{sec:color:local}

Levin~\etal\cite{closedformpami} proposed the use of the gradient of the alpha channel as the amount of local information flow for the problem of layer color estimation.
They solely rely on this form of information flow for propagating the colors.
This local information flow basically enforces neighboring pixels to have similar colors if there is an alpha transition.
This flow, which we refer to as \mbox{\emph{$\alpha$-transition flow}}, can be represented by the following energy:
\begin{equation}
E_{\nabla\tilde\alpha} = \sum_{\forall p} \sum_{q \in \mathcal{N}_p^L}
        |\nabla\tilde\alpha_{(p-q)}| \left( \left( f_p - f_q  \right)^2 + \left( b_p - b_q  \right)^2  \right),
\label{eq:alphagradientflow}
\end{equation}
where $\nabla\tilde\alpha$ represents the alpha gradient.
We compute the gradients in the image plane using the 3-tap separable filters of Farid and Simoncelli~\cite{faridsimoncelli}.
Note that the neighborhood is defined as the local $3\times3$ neighborhood similar to the local information flow in Section~\ref{sec:local}.

The transition flow helps around small regions with alpha gradient but does not propagate information in flat-alpha regions, such as pure foreground or background regions or regions with flat opacity.
We propose a new smoothness measure to address this issue, which we call \mbox{\emph{no-transition flow}}.
The no-transition flow enforces spatial smoothness in regions with small color and alpha gradients:
\begin{equation}
E_{\nabla\ve{c}\tilde\alpha} = 
\sum_{\forall p} \sum_{q \in \mathcal{N}_p^L}
w_{p,q}^{\nabla\ve{c}\tilde\alpha}
\left( \left( f_p - f_q  \right)^2 + \left( b_p - b_q  \right)^2  \right)
\label{eq:alphacolorgradientflow}
\end{equation}
where $w_{p,q}^{\nabla\ve{c}\tilde\alpha} = \left(1 - |\nabla\tilde\alpha_{(p-q)}|\right)\left(1 - ||\nabla\ve{c}_{(p-q)}||\right)$ and $||\nabla\ve{c}_{(p-q)}||$ is the $L_2$ norm of the vector formed by gradients of the individual color channels.
This term increases the performance around slow alpha transitions and flat-alpha regions, as well as around sharp color edges in the image.

No-transition flow already improves the performance quite noticably as seen in Figure~\ref{fig:stepByStep_color}(b).
However, using only local information flows perform poorly in remote areas such as the end of long hair filaments (Figure~\ref{fig:groundColorComp}(a)) or isolated areas (Figure~\ref{fig:stepByStep_color}, bottom inset).
In order to increase the performance in these type of challenging areas, we make use of two types of non-local information flows.


\subsubsection{Color-mixture information flow}
\label{sec:color:mixture}

The basic principle of color mixture as introduced in Section~\ref{sec:mixture} also applies to the relationship between layer colors of pixels in the same neighborhood --- if we represent the color and alpha of a pixel as a weighted combination of the colors and alpha of several others, those weights should also represent the layer color relation between the pixels.
Since we have $\tilde\alpha$'s as additional information in the layer color estimation scenario, we extend the formulation of color-mixture flow to better fit the layer color estimation problem.
Similar to its use in alpha estimation, it provides a well-connected graph and allows dense share of information.
The performance improvement by the introduction of the color-mixture energy can be seen in Figure~\ref{fig:stepByStep_color}(c).

In the layer color estimation scenario, we optimize for both foreground and background colors in the same formulation.
It should be emphasized that, as it is apparent from \refeq{eq:compositing}, the foreground and background colors are undefined for regions with $\tilde\alpha = 0$ and $\tilde\alpha = 1$, respectively.
This requires us to avoid using color-mixture flow into $\tilde\unm$ from $\tilde\bgm$ for $\ve{f}$ and from $\tilde\fgm$ for $\ve{b}$.
We address this by defining two different neighborhoods and computing individual color-mixture flows for $\ve{f}$ and $\ve{b}$.

For $\ve{f}$, we define the neighborhood $\mathcal{N}_p^{\tilde\unm\tilde\fgm}$ by finding $K_{CM}$ nearest neighbors in $(\tilde\unm \cup \tilde\fgm)$ using the feature vector $[r,g,b,\tilde\alpha,\tilde{x},\tilde{y}]^T$.
We then compute the weights $w_{p,q}^{C\tilde\fgm}$ as
\begin{equation}
\argmin_{w_{p,q}^{C\tilde\fgm}}
\left\| 
\begin{bmatrix}
\ve{c}_p \\ \tilde\alpha_p
\end{bmatrix}
 - \sum_{q\in\mathcal{N}_p^{CM}} w_{p,q}^{C\tilde\fgm} 
 \begin{bmatrix}
\ve{c}_q \\ \tilde\alpha_q
\end{bmatrix} 
\right\| ^ 2.
\label{eq:lleColor}
\end{equation}
Notice that the search space and the weight computation includes $\tilde\alpha$ in addition to the color and location of pixels.

We compute the background conjugates of the neighborhood and weights, $\mathcal{N}_p^{\tilde\unm\tilde\bgm}$ and $w_{p,q}^{C\tilde\bgm}$, in the same way, and define our color-mixture energy for layer color estimation:
%
\begin{equation*}
E_{CM}^{fb} =
\sum_{p \in \tilde\unm}
(( f_p - \sum_{q \in \mathcal{N}_p^{\tilde\unm\tilde\fgm}} w^{C\tilde\fgm}_{p,q} f_q )^2 +
( b_p - \sum_{q \in \mathcal{N}_p^{\tilde\unm\tilde\bgm}} w^{C\tilde\bgm}_{p,q} b_q )^2).
\label{eq:colormixtureenergyforcolor}
\end{equation*}
%


\subsubsection{Intra-$\tilde\unm$ information flow}
\label{sec:color:insideUnknown}

Intra-\unt\ information flow, as detailed in Section~\ref{sec:insideUnknown}, distributes the information between similar-colored pixels inside the unknown region without giving spatial proximity too much emphasis.
Its behaviour is also very useful in the case of color estimation, as it makes the foreground colors more coherent throughout the image.
For example, in Figure~\ref{fig:stepByStep_color}, bottom inset shows that the addition of intra-\unt\ flow helps in getting a more realistic color to the isolated plastic region between the two black lines.

We make modifications to intra-\unt\ flow similar to the modifications we made to color-mixture flow, in order to make use of the available information coming form $\tilde\alpha$'s.
 
 We find $K_\unm$ nearest neighbors only inside $\tilde\unm$ to determine $\mathcal{\hat{N}}_p^{\tilde\unm}$ using the feature vector defined as $\ve{v}^{\ve{c}} = [r, g, b, \tilde\alpha, \tilde{x}/20, \tilde{y}/20]^T$. We then determine the amount of information flow between two non-local neighbors as:
\begin{equation}
w_{p,q}^{\tilde\unm} = \max\left(1 - \left\| \ve{v}_p^{\ve{c}} - \ve{v}_q^{\ve{c}} \right\|_1,\ 0\right) \quad \forall q \in \mathcal{N}_p^{\tilde\unm}.
\label{eq:knnweightcolor}
\end{equation}
With the weights determined, we can define the energy function representing the intra-$\tilde\unm$ flow:
\begin{equation}
E_{\tilde\unm\tilde\unm} = \sum_{p \in \tilde\unm} \sum_{q \in \mathcal{N}_p^{\tilde\unm}} w^{\tilde\unm}_{p,q} \left(\left( f_p - f_q \right)^2 + \left( b_p - b_q \right)^2 \right).
\label{eq:similarityEnergyColor}
\end{equation}

Note that in the color estimation formulation, we exclude the \mbox{\knt-to-\unt} information flow because we observed that the adaptation of the method in Section~\ref{sec:knownToUnknown} to color estimation does not improve the quality of the final result.


\subsection{Linear system and energy minimization}
\label{sec:color:linearSystem}

The final energy function for layer color estimation is the combination of the four types of information flow defined in Sections~\ref{sec:color:local}~to~\ref{sec:color:insideUnknown}:
\begin{equation}
E_{\ve{c}} = \sigma_L E_{\nabla\alpha} + \sigma_L E_{\nabla\ve{c}\alpha} + E^{fb}_{CM} + \sigma_{\unm\unm} E_{\tilde\unm\tilde\unm} + \lambda E_{\text{COMP}},
\end{equation}
where $\sigma_L$, $\sigma_{\unm\unm}$ and $\lambda$ are defined in Section~\ref{sec:linearSystem} and $E_{\text{COMP}}$ represents the deviation from the compositing equation constraint:
\begin{equation}
E_{\text{COMP}} = \sum_{\forall p} \left( c_p - \alpha^I_p f - (1 - \alpha^I_p) b \right)^2.
\label{eq:compositingDeviation}
\end{equation}
$E_{\ve{c}}$ is defined and minimized independently for each color channel.

Following the same strategy as we did in Section~\ref{sec:linearSystem}, we rewrite the energy function $E_{\ve{c}}$ in the matrix form, this time as a $2N \times 2N$ linear system, and solve it for foreground and background colors for 3 times, once for each color channel, using the preconditioned conjugate gradients method~\cite{pcg}.

%

\section{Results and discussion}
\label{sec:results}

\begin{table*}[t]
\caption{
Our scores in the alpha matting benchmark~\cite{alphabenchmark} together with the top-performing published methods at the time of submission.
\emph{S}, \emph{L} and \emph{U} denote the three trimap types, small, large and user, included in the benchmark.
Bold and blue numbers represent the best scores in the benchmark.
}
\resizebox{\linewidth}{!}
{
{\setlength{\extrarowheight}{4pt}
\setlength\arrayrulewidth{1pt}
\rowcolors{2}{gray!25}{}
\begin{tabular}{
     !{\color{gray!25}\vrule} 
     l 
     !{\color{gray!25}\vrule}
     c
     !{\color{gray!25}\vrule} 
     c c c 
     !{\color{gray!25}\vrule} 
     c c c 
     !{\color{gray!25}\vrule} 
     c c c 
     !{\color{gray!25}\vrule} 
     c c c 
     !{\color{gray!25}\vrule} 
     c c c 
     !{\color{gray!25}\vrule} 
     c c c 
     !{\color{gray!25}\vrule} 
     c c c 
     !{\color{gray!25}\vrule} 
     c c c 
     !{\color{gray!25}\vrule} 
     c c c 
     !{\color{gray!25}\vrule}
     }
    \arrayrulecolor{gray!25}\hline
    &
    \multicolumn{4}{c !{\color{gray!25}\vrule} }{Average Rank} & 
    \multicolumn{3}{c !{\color{gray!25}\vrule} }{Troll} &
    \multicolumn{3}{c !{\color{gray!25}\vrule} }{Doll} &
    \multicolumn{3}{c !{\color{gray!25}\vrule} }{Donkey} &
    \multicolumn{3}{c !{\color{gray!25}\vrule} }{Elephant} &
    \multicolumn{3}{c !{\color{gray!25}\vrule} }{Plant} &
    \multicolumn{3}{c !{\color{gray!25}\vrule} }{Pineapple} &
    \multicolumn{3}{c !{\color{gray!25}\vrule} }{Plastic bag} &
    \multicolumn{3}{c !{\color{gray!25}\vrule} }{Net}
    \\
    & 
    \emph{Overall} & 
    \emph{S} &     \emph{L} &     \emph{U} &
    \emph{S} &     \emph{L} &     \emph{U} &
    \emph{S} &     \emph{L} &     \emph{U} &
    \emph{S} &     \emph{L} &     \emph{U} &
    \emph{S} &     \emph{L} &     \emph{U} &
    \emph{S} &     \emph{L} &     \emph{U} &
    \emph{S} &     \emph{L} &     \emph{U} &
    \emph{S} &     \emph{L} &     \emph{U} &
    \emph{S} &     \emph{L} &     \emph{U}
    \\
    \arrayrulecolor{gray!25}\hline
 \multicolumn{29}{!{\color{gray!25}\vrule}c !{\color{gray!25}\vrule} }{Sum of Absolute Differences}\\
    \arrayrulecolor{gray!25}\hline
    Ours & 2.7 & 3.3 & 2.3 & 2.6 & 
    \topscore{10.3}  & \topscore{11.2}  & {12.5}  & 
    {5.6}  & {7.3}  & {7.3}  & 
    {3.8}  & {4.1}  & {3}  & 
    {1.4}  & {2.3}  & {2.0}  & 
    {5.9}  & {7.1}  & \topscore{8.6}  & 
    {3.6}  & {5.7}  & {4.6}  & 
    {18.3}  & {19.3}  & \topscore{15.8}  & 
    {20.2}  & {22.2}  & {22.3}
    \\
    DIM \cite{deepmatting} & 2.9 & 3.6 & 2.3 & 2.8 & 
    {10.7}  & {11.2}  & \topscore{11.0}  & 
    \topscore{4.8}  & {5.8}  & \topscore{5.6}  & 
    \topscore{2.8}  & \topscore{2.9}  & \topscore{2.9}  & 
    \topscore{1.1}  & \topscore{1.1}  & \topscore{2.0}  & 
    {6.0}  & {7.1}  & {8.9}  & 
    \topscore{2.7}  & \topscore{3.2}  & \topscore{3.9}  & 
    {19.2}  & {19.6}  & {18.7}  & 
    {21.8}  & {23.9}  & {24.1}
    \\
    DCNN \cite{dcnn} &  4.0 & 5.4 & 2.3 & 4.3 & 
    {12.0}  & {14.1}  & {14.5}  & 
    {5.3}  & {6.4}  & {6.8}  & 
    {3.9}  & {4.5}  & {3.4}  & 
    {1.6}  & {2.5}  & {2.2}  & 
    {6.0}  & \topscore{6.9}  & {9.1}  & 
    {4.0}  & {6.0}  & {5.3}  & 
    {19.9}  & \topscore{19.2}  & {19.1}  & 
    \topscore{19.4}  & \topscore{20.0}  & \topscore{21.2}
    \\
    CSC \cite{csc} &   11 & 14.4 & 7.4 & 11.3 & 
    13.6 & 15.6 & 14.5 & 6.2 & 7.5 & 8.1 & 4.6 & 4.8 & 4.2 & 1.8 & 2.7 & 2.5 & 5.5 & 7.3 & 9.7 & 4.6 & 7.6 & 6.9 & 23.7 & 23.0 & 21.0 & 26.3 & 27.2 & 25.2
    \\
    \arrayrulecolor{gray!25}\hline
 \multicolumn{29}{!{\color{gray!25}\vrule}c !{\color{gray!25}\vrule} }{Mean Squared Error}\\
    \arrayrulecolor{gray!25}\hline
    Ours & 
    4.0 &
    5.4 & 2.8 & 3.8 & 
    \topscore{0.3} & \topscore{0.4} & {0.5} & 
    {0.3} & {0.4} & {0.5} & 
    {0.3} & {0.3} & {0.2} & 
    {0.1} & {0.1} & {0.1} & 
    {0.4} & {0.4} & \topscore{0.6} & 
    {0.2} & {0.3} & {0.3} & 
    {1.3} & {1.2} & \topscore{0.8} & 
    {0.8} & {0.8} & {0.9}
    \\
    DCNN \cite{dcnn} &  
    4.3 &
    5.3 & 2.5 & 5.0 & 
    {0.4} & {0.5} & {0.7} & 
    {0.2} & \topscore{0.3} & {0.4} & 
    {0.2} & {0.3} & {0.2} & 
    {0.1} & {0.1} & \topscore{0.1} & 
    {0.4} & \topscore{0.4} & {0.8} & 
    {0.2} & {0.4} & {0.3} & 
    {1.3} & {1.2} & {1.0} & 
    \topscore{0.7} & \topscore{0.7} & {0.9}
    \\
    DIM \cite{deepmatting} & 
    4.6 &
    3.5 & 4.0 & 6.3 & 
    {0.4} & {0.4} & \topscore{0.4} & 
    \topscore{0.2} & {0.3} & \topscore{0.3} & 
    \topscore{0.1} & \topscore{0.1} & \topscore{0.2} & 
    \topscore{0} & \topscore{0} & {0.2} & 
    {0.5} & {0.6} & {1} & 
    \topscore{0.2} & \topscore{0.2} & {0.4} & 
    \topscore{1.1} & \topscore{1.1} & {1.1} & 
    {0.8} & {0.9} & {1}
    \\
    LNSP \cite{lnsp} & 
    10.2 &
    7.6 & 9.6 & 13.3 &
    {0.5} & {1.9} & {1.2} & 
    {0.2} & {0.4} & {0.5} & 
    {0.3} & {0.4} & {0.2} & 
    \topscore{0.0} & {0.1} & {0.2} & 
    {0.4} & {0.5} & {0.8} & 
    \topscore{0.2} & {0.3} & {0.4} & 
    {1.4} & {1.2} & {0.8} & 
    {1.0} & {1.1} & {1.5}
    \\
    \arrayrulecolor{gray!25}\hline
\end{tabular}
}
}
\label{tab:benchmark}
\end{table*}

We evaluate the proposed methods for matting, matte regularization, layer color estimation and green-screen keying with comparisons to the state-of-the-art of each application.

\subsection{Matte estimation}
\label{sec:results:matting}

\begin{figure*}
\resizebox{\linewidth}{!}
{
\centering
{
\setlength{\extrarowheight}{-10pt}
\begin{tabular}{c | c}
\arrayrulecolor{gray!50}
    \showimagew[\linewidth]{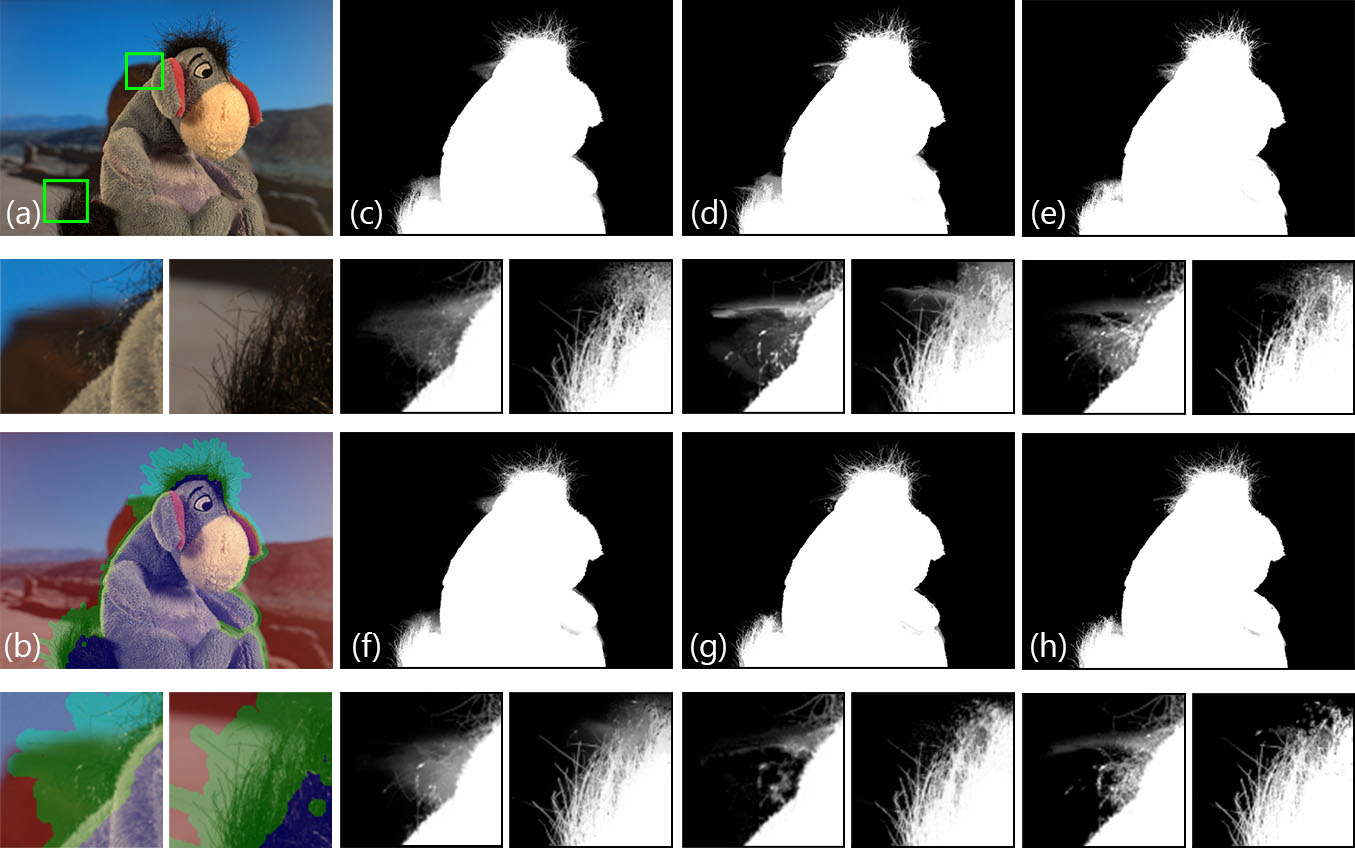} & 
    \showimagew[\linewidth]{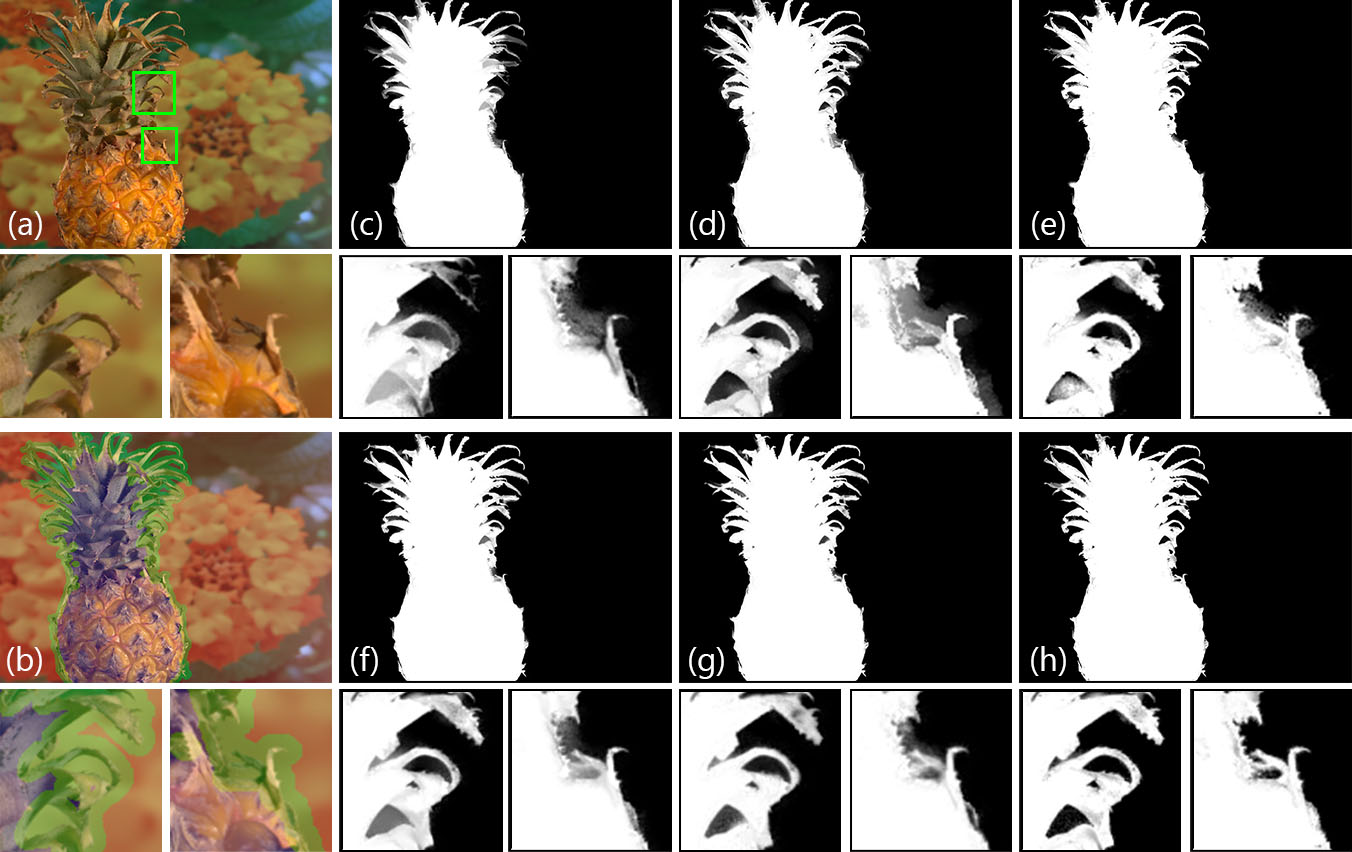}\\
     \hline \\
    \showimagew[\linewidth]{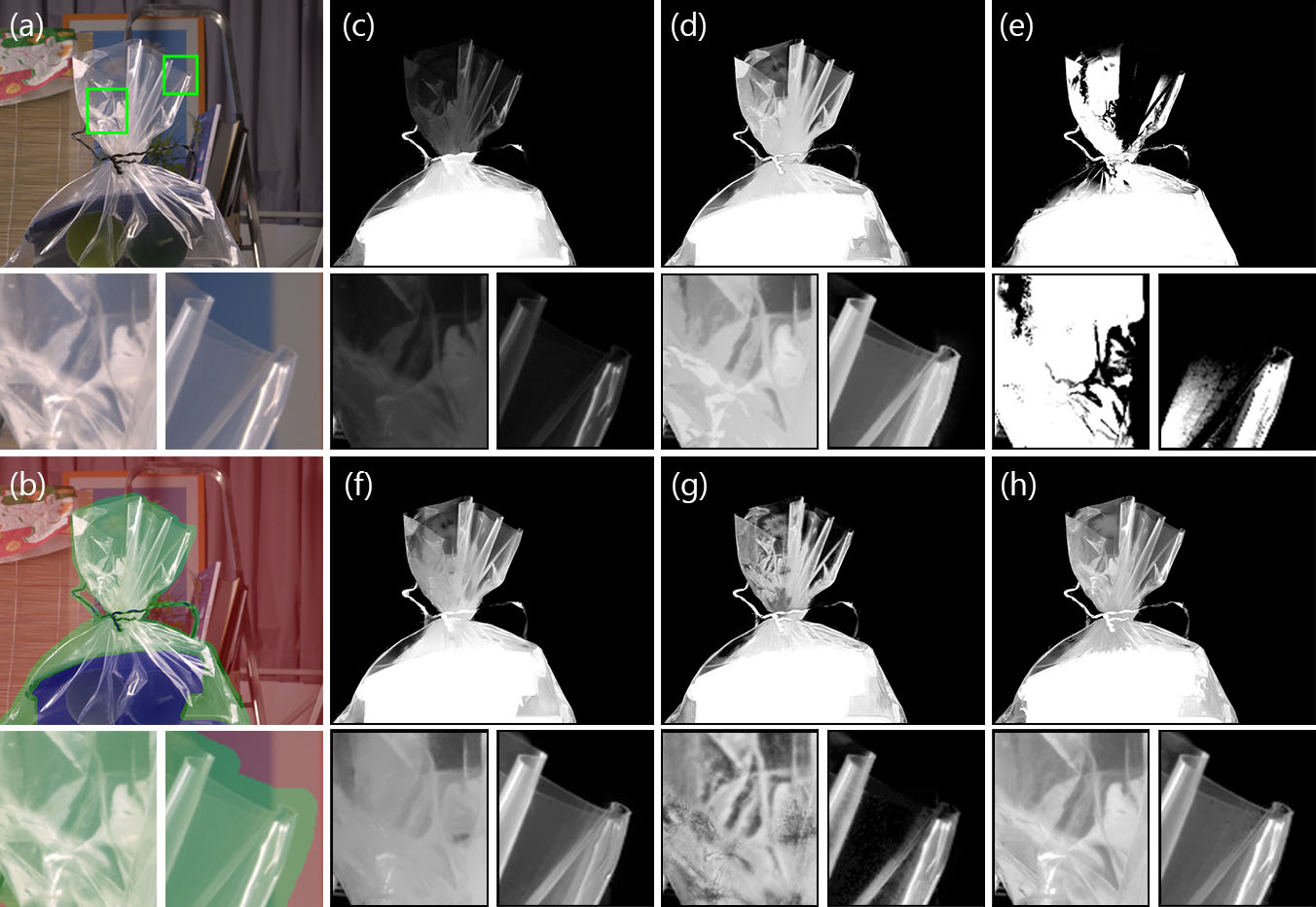} & 
    \showimagew[\linewidth]{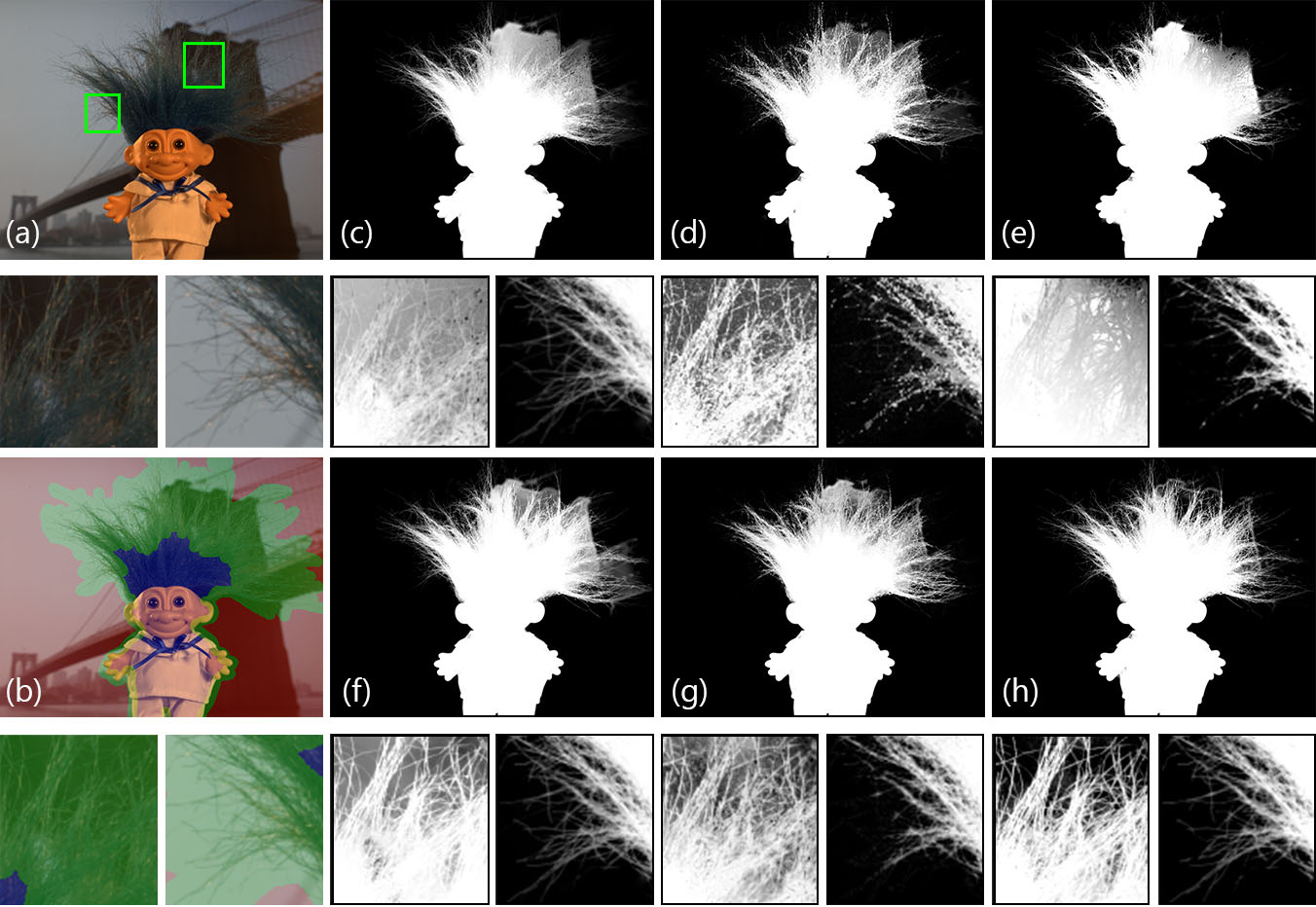}\\
\end{tabular}
}
}
\caption{
Several examples from the alpha matting benchmark~\cite{alphabenchmark} are shown (a) with trimaps overlayed onto the images (b).
The mattes are computed by 
closed-form matting~\cite{closedformpami} (c), 
KNN matting~\cite{knnpami} (d), 
manifold-preserving edit propagation~\cite{manifold} (e), 
LNSP matting~\cite{lnsp} (f), 
DCNN matting~\cite{dcnn} (g) and
the proposed method (h).
See text for discussion.
}
\negvspace
\negvspace
\label{fig:LargeComparison}
\end{figure*}


We quantitatively evaluate the proposed algorithm using the public alpha matting benchmark~\cite{alphabenchmark} in Table~\ref{tab:benchmark}.
At the time of submission, our method ranks in the first place according to the sum-of-absolute-differences (SAD) and mean-squared error (MSE) metrics.
Our proof-of-concept implementation in Matlab requires on average 50 seconds to process a benchmark image.

Our performance in the test set by Xu~\etal\cite{deepmatting} is shown in Table~\ref{tab:dimscores}. 
This test set of 1000 images accompany
\begin{wraptable}{r}{0.5\linewidth}
\caption{
    Matting performance on the test set of DIM~\cite{deepmatting}.
}
\resizebox{\linewidth}{!}
{
{\setlength{\extrarowheight}{4pt}
\setlength\arrayrulewidth{1pt}
\rowcolors{2}{gray!25}{}
\begin{tabular}{
!{\color{gray!25}\vrule} 
l
!{\color{gray!25}\vrule} 
c 
!{\color{gray!25}\vrule} 
!{\color{gray!25}\vrule} 
c 
!{\color{gray!25}\vrule} 
}
    \arrayrulecolor{gray!25}\hline
    &
    {SAD} & 
    {MSE}
    \\
    DIM~\cite{deepmatting} & 
    50.4 &
    0.014 
    \\
    Ours & 
    100.6 & 
    0.038 
    \\
    DCNN~\cite{dcnn} & 
    161.4 &
    0.087 
    \\
    CF~\cite{closedformpami} & 
    168.1 &
    0.091 
    \\
    KNN~\cite{knnpami} & 
    175.4 &
    0.103 
    \\
    \arrayrulecolor{gray!25}\hline
\end{tabular}
}
}
\label{tab:dimscores}
\end{wraptable}\setcounter{table}{2}

a data-driven approach to matting.
One advantage of using a deep network for this problem, such as DIM~\cite{deepmatting}, is that the network can infer the matte even when there is no foreground region defined in the trimap due to heavy transparency, and their test set includes several such examples.
Affinity-based and sampling-based approaches, however, assume both known regions are present when they are modeling the color models of affinities.
While this can be seen as a shortcoming, the images without well-defined regions inadvertently skew the scores in this dataset.
We perform better than competing methods except for DIM in this dataset, and our scores improve to be 76.5 (SAD) and 0.021 (MSE) when the images that violate our assumptions are removed.

We also compare our results qualitatively with the closely related methods in \reffig{fig:LargeComparison}.
We use the results that are available on the matting benchmark for all except manifold-preserving matting~\cite{manifold} which we implemented ourselves. 
\reffig{fig:LargeComparison}(c,d,e) show that using only one form of information flow is not effective in a number of scenarios such as wide unknown regions or holes in the foreground.
The strategy DCNN matting~\cite{dcnn} follows is using the results of closed-form and KNN matting directly rather than formulating a combined energy using their affinity definitions.
When both methods fail, the resulting combination also suffers from the errors as it is apparent in the pineapple and troll examples.
The neural network they propose also seems to produce mattes that appear slightly blurred.
LNSP matting~\cite{lnsp}, on the other hand, has issues around regions with holes (pineapple example) or when the foreground and background colors are similar (donkey and troll examples).
It can also oversmooth some regions if the true foreground colors are missing in the trimap (plastic bag example).
Our method performs well in these challenging scenarios mostly because 
of the \mbox{intra-unknown} and \mbox{unknown-to-known} connections which results in a more robust linear system.

\begin{figure*}
{
\footnotesize
\begin{tabular}
{   K{0.225\linewidth}
    K{0.225\linewidth}
    K{0.225\linewidth}
    K{0.225\linewidth}
}
Input and ground-truth&
Regularization of KL-D~\cite{kldiv}&
Regularization of SM~\cite{shared}&
Regularization of CS~\cite{comprehensive}
\end{tabular}
}
\showimagew[\linewidth]{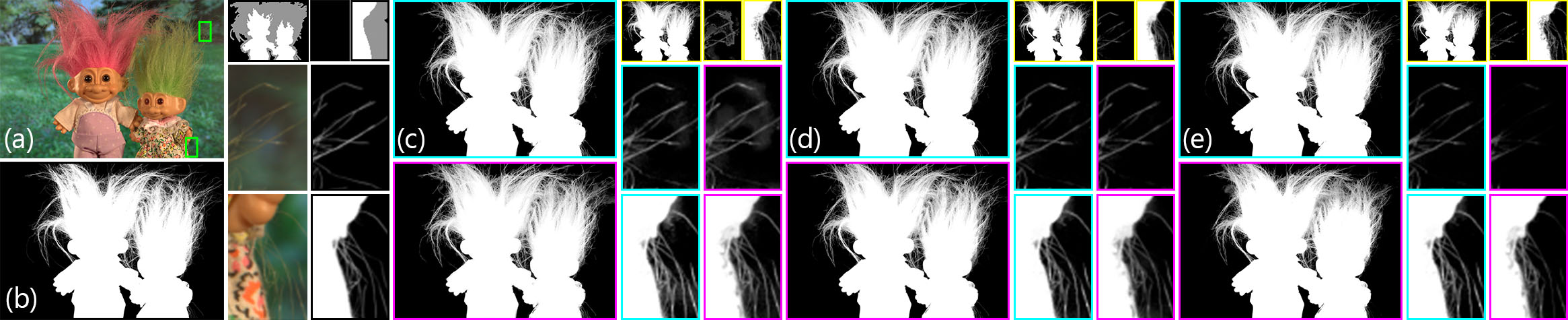}
\caption{
Matte regularization using the proposed method (cyan) or \cite{shared} (magenta) for three sampling-based methods (yellow).
Our method is able to preserve remote details while producing a clean matte (top inset) and preserve sharpness even around textured areas (bottom).
}
\negvspace
\negvspace
\label{fig:MatteRefinementLargeComparison}
\end{figure*}

We evaluate the sensitivity of our method against different parameter values on the training dataset of the matting benchmark~\cite{alphabenchmark}.
Table~\ref{tab:params} shows that different values for the parameters generally have only a small effect on the performance on average.

\begin{table}[t]
\caption{
    Average percentage performance change with changing parameters using 27 images and 2 trimaps from the benchmark.
}
\resizebox{\linewidth}{!}
{
{\setlength{\extrarowheight}{4pt}
\setlength\arrayrulewidth{1pt}
\rowcolors{2}{gray!25}{}
\begin{tabular}{
!{\color{gray!25}\vrule} 
l
!{\color{gray!25}\vrule} 
c 
!{\color{gray!25}\vrule}
!{\color{gray!25}\vrule}
c 
!{\color{gray!25}\vrule}
c
!{\color{gray!25}\vrule} 
!{\color{gray!25}\vrule}
c 
!{\color{gray!25}\vrule}
c
!{\color{gray!25}\vrule} 
!{\color{gray!25}\vrule}
c 
!{\color{gray!25}\vrule}
c
!{\color{gray!25}\vrule} 
!{\color{gray!25}\vrule}
c 
!{\color{gray!25}\vrule}
c
!{\color{gray!25}\vrule}
}
    \arrayrulecolor{gray!25}\hline
    Param. & 
    Def. & 
    Val. &
    Perf. &
    Val. &
    Perf. &
    Val. &
    Perf. &
    Val. &
    Perf.
    \\
    \arrayrulecolor{gray!25}\hline
    \arrayrulecolor{gray!25}\hline
    $K_{CM}$ & 
    20 & 
    10 &
    1.07 \% &
    15 &
    0.44 \% &
    25 &
    -0.46 \% &
    30 &
    -0.62 \%
    \\
    $K_{\knm-\unm}$ & 
    7 & 
    1 &
    -0.83 \% &
    4 &
    -0.41 \% &
    10 &
    0.12 \% &
    13 &
    0.22 \%
    \\
    $K_{\unm-\unm}$ & 
    5 & 
    1 &
    -0.15 \% &
    3 &
    -0.1 \% &
    7 &
    0.08 \% &
    9 &
    0.11 \%
    \\
    $\sigma_{\knm-\unm}$ & 
    0.05 & 
    0.01 &
    -6.44 \% &
    0.025 &
    -2.1 \% &
    0.075 &
    0.66 \% &
    0.09 &
    0.87 \%
    \\
    $\sigma_{\unm-\unm}$ & 
    0.01 & 
    0.001 &
    -0.7 \% &
    0.005 &
    -0.1 \% &
    0.02 &
    -0.47 \% &
    0.05 &
    -3.12 \%
    \\
    \arrayrulecolor{gray!25}\hline
\end{tabular}
}
}
\negvspace
\negvspace
\label{tab:params}
\end{table}

\subsection{Matte regularization}
\label{sec:results:regularization}


We also compare the proposed post-processing method detailed in Section~\ref{sec:postprocessing} with the state-of-the-art method by Gastal and Oliveira~\cite{shared} on the training dataset provided by Rhemann~\etal\cite{alphabenchmark}.
We computed the non-smooth alpha values and confidences using the publicly available source code for comprehensive sampling~\cite{comprehensive}, KL-divergence sampling~\cite{kldiv} and shared matting~\cite{shared}.
Table~\ref{tab:refinement} shows the percentage improvement we achieve over Gastal and Oliveira~\cite{shared} for each algorithm using SAD and MSE as error measures.
\reffig{fig:MatteRefinementLargeComparison} shows an example for regularizing all three sampling-based methods.
As the information coming from alpha values and their confidences found by the sampling-based method is distributed more effectively by the proposed method, the challenging regions such as fine structures or holes detected by the sampling-based method are preserved when our method is used for post-processing.

\begin{figure*}
{
    \footnotesize
    \begin{tabular}
    {   K{0.130\linewidth}
        K{0.190\linewidth}
        K{0.190\linewidth}
        K{0.190\linewidth}
        K{0.190\linewidth}
    }
    Input image&
    Ground truth&
    Closed-form colors&
    KNN colors&
    Ours
    \end{tabular}
}
\showimagew[\linewidth]{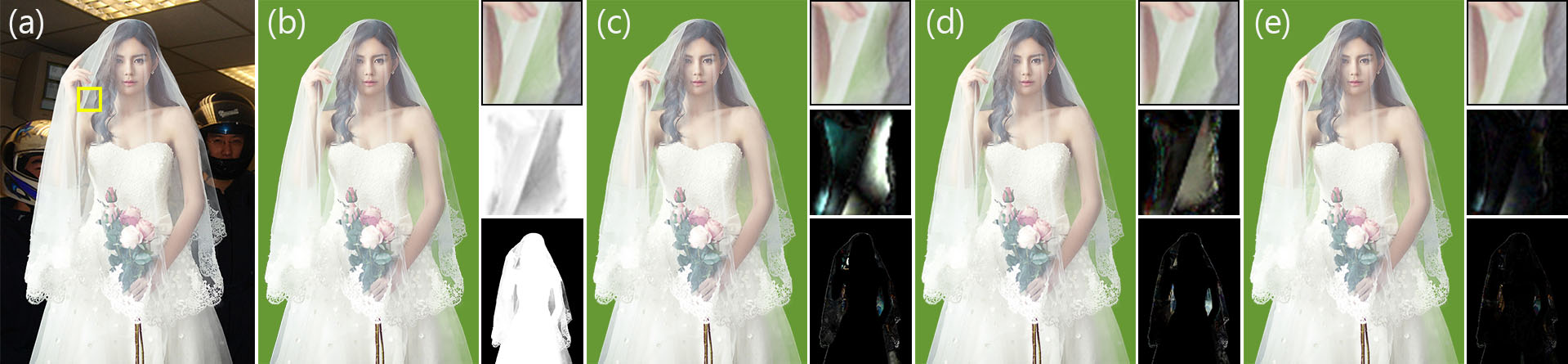}\\
\showimagew[\linewidth]{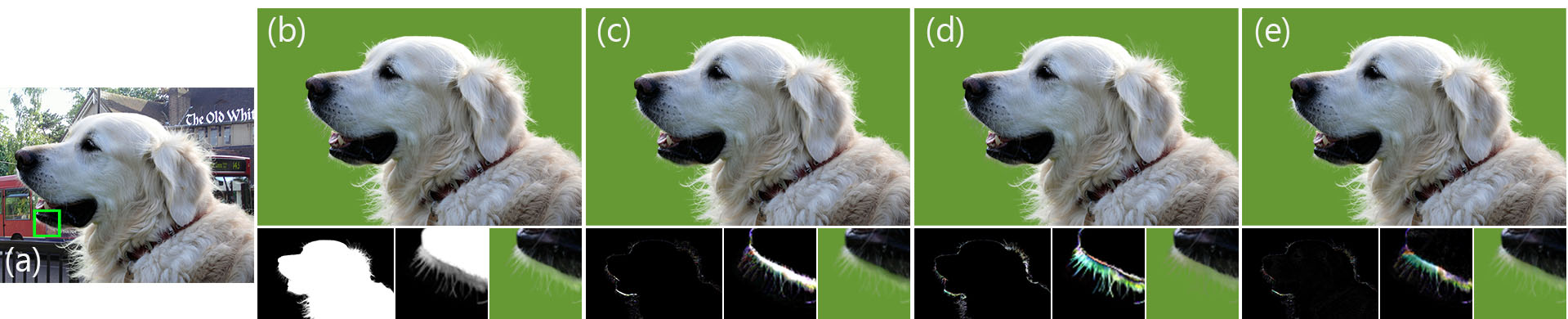}\negvspace
\caption{
Color estimation results of three algorithms together with the ground truth colors and matte (b).
The bottom-right in each set shows per-pixel absolute difference between the estimation and ground truth multiplied by ten.
See text for discussion.
\negvspace
}
\label{fig:groundColorComp}
\end{figure*}

\subsection{Layer color estimation}
\label{sec:results:color}
We evaluate our layer color estimation method against the closed-form color estimation~\cite{closedformpami} and KNN colors~\cite{knnpami},
\begin{wraptable}{r}{0.5\linewidth}
    \caption{
        Layer color estimation performance on the test set of DIM~\cite{deepmatting}.
    }
    \resizebox{\linewidth}{!}
    {
    {\setlength{\extrarowheight}{4pt}
    \setlength\arrayrulewidth{1pt}
    \rowcolors{2}{gray!25}{}
    \begin{tabular}{
    !{\color{gray!25}\vrule} 
    l
    !{\color{gray!25}\vrule} 
    c
    !{\color{gray!25}\vrule} 
    c
    !{\color{gray!25}\vrule} 
    }
        \arrayrulecolor{gray!25}\hline
        &
        {SAD} & 
{MSE}
        \\
        Ours & 
        $3.8\times10^{3}$ & 
        $6.9\times10^{-4}$
        \\
        CF~\cite{closedformpami} & 
        $4.3\times10^{3}$ & 
        $9.2\times10^{-4}$
        \\
        KNN~\cite{knnpami} & 
        $4.7\times10^{3}$ & 
        $8.4\times10^{-4}$
        \\
        \arrayrulecolor{gray!25}\hline
    \end{tabular}
    }
    }
    \label{tab:colorEst}
    \end{wraptable}

on the test set of deep image matting~\cite{deepmatting} using the ground-truth alphas as input.
Closed-form colors only use a single local affinity to propagate the colors from the foreground, and this creates artifacts around holes in the foreground (Figure~\ref{fig:groundColorComp}, top) or incorrect colors being propagated to nearby regions (bottom).
KNN colors, on the other hand, uses only the similarity affinity and it typically generates flat-colored regions, which results in erroneous values especially around hair and fur.
Our multi-affinity approach is able to correctly estimate the colors even in the isolated regions or intricate structures.
These properties are also reflected in the quantitative comparison, as shown in Table~\ref{tab:colorEst}.

\begin{table}[t]
\caption{
    Performance improvement achieved when our matte regularization method replaces~\cite{shared} in the post-processing steps of 3 sampling-based methods.
    The training dataset in \cite{alphabenchmark} was used for this experiment.
}
\resizebox{\linewidth}{!}
{
{\setlength{\extrarowheight}{4pt}
\setlength\arrayrulewidth{1pt}
\rowcolors{2}{gray!25}{}
\begin{tabular}{
!{\color{gray!25}\vrule} 
l
!{\color{gray!25}\vrule} 
c 
!{\color{gray!25}\vrule}
c c 
!{\color{gray!25}\vrule} 
!{\color{gray!25}\vrule} 
c 
!{\color{gray!25}\vrule}
c c
!{\color{gray!25}\vrule} 
}
    \arrayrulecolor{gray!25}\hline
    &
    \multicolumn{3}{ c !{\color{gray!25}\vrule} !{\color{gray!25}\vrule} }{Sum of Absolute Differences} & 
    \multicolumn{3}{ c !{\color{gray!25}\vrule} }{Mean Squared Error}
    \\
    \arrayrulecolor{gray!25}\hline
    &
    Overall &
    S &
    L &
    Overall &
    S &
    L
    \\
    KL-D~\cite{kldiv} & 
    24.4 \% &
    22.4 \% &
    26.5 \% &
    28.5 \% &
    25.9 \% &
    31.0 \% 
    \\
    SM~\cite{shared} & 
    6.0 \% &
    3.7 \% &
    8.4 \% &
    13.6 \% &
    8.5 \% &
    18.8 \% 
    \\
    CS~\cite{comprehensive} & 
    4.9 \% &
    10.0 \% &
    -0.1 \% &
    18.7 \% &
    25.5 \% &
    11.8 \% 
    \\
    \arrayrulecolor{gray!25}\hline
\end{tabular}
}
}
\label{tab:refinement}
\end{table}


\begin{figure*}[t]
\showimagew[\linewidth]{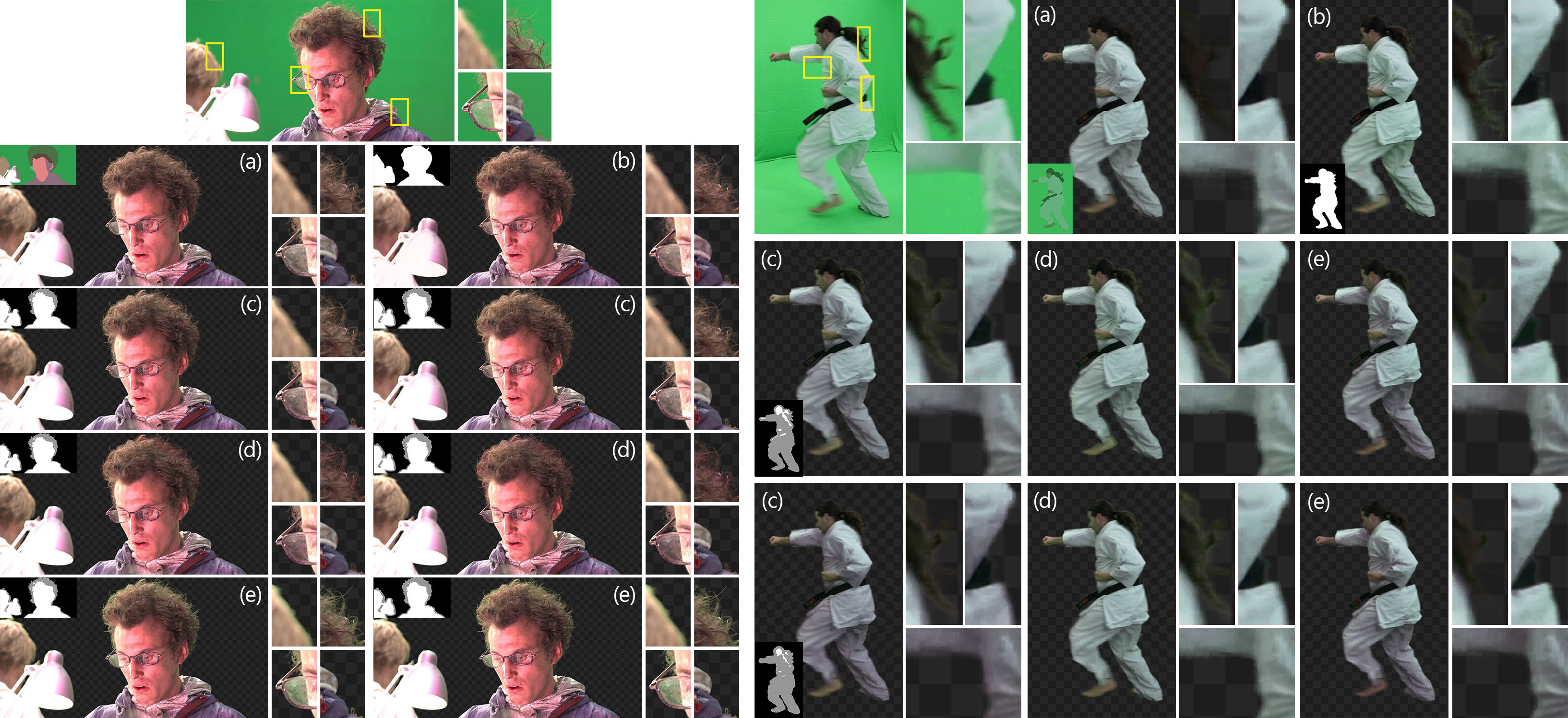}
\caption{
Green-screen keying results of GSK~\cite{keying} with its input called
\emph{local color models} (a) and of SCS~\cite{softcolorsegm} with the mask
needed for a clean result (b) together with the proposed method (c),
comprehensive sampling~\cite{comprehensive} (d) and KNN matting~\cite{knnpami} (e)
using two trimaps, one narrow and one wide, for each example.
See text for discussion.
}
\label{fig:keying}
\end{figure*}

\subsection{Green-screen keying}
\label{sec:results:keying}

Green-screen keying is a more constrained version of the natural image matting problem in which the background is mostly of single color.
Despite the more constrained setup, it is challenging to get clean foregrounds for compositing.
Aksoy~\etal\cite{keying} show that common natural matting algorithms fail to get satisfactory results despite their performance on the matting benchmark.

We compare the performance of our method to that of the interactive green-screen keying method by Aksoy~\etal\cite{keying} (GSK) and unmixing-based soft color segmentation~\cite{softcolorsegm} (SCS) as well as KNN matting~\cite{knnpami} and comprehensive sampling~\cite{comprehensive} in Figure~\ref{fig:keying}.
GSK requires local color models, a subset of entries in their color model, and SCS requires a binary map to clean the noise in the background.
The matting methods including ours require trimaps and we show results for two trimaps used for comparisons in~\cite{keying}.
We computed the foreground colors for our method and comprehensive sampling using our color estimation method, and KNN colors for KNN matting.
We observed that the choice of color estimation method does not change the typical artifacts we see in KNN matting and comprehensive sampling.
GSK and SCS compute foreground colors together with the alpha values.


Top example in Figure~\ref{fig:keying} shows that KNN matting overestimates alpha values in critical areas and this results in a green halo around the foreground.
In contrast, we see a reddish hue in the hair and around the glasses for comprehensive sampling.
This is due to the underestimation of alpha values in those areas.
The bottom example shows that both competing matting methods fail to get rid of the color spill, \ie indirect illumination from the background.
The proposed method successfully extracts the foreground matte and colors in both challenging cases and gives comparable results to the state-of-the-art in green-screen keying.
It can also be seen that the effect of different trimaps is minimal in both cases.
A successful matting approach requires less input than GSK (the local color models are conceptually similar to a multi-channel trimap and requires more time to generate than a trimap) and is robust against color spill unlike SCS, which makes our method a viable option for green-screen keying.

Although the images shown in Figure~\ref{fig:keying} have the resolution of 1080p, the average time our matte estimation was around 20 seconds, which is lower than our average for the matting benchmark.
The reason is that the time required to construct and solve our linear system mostly depends on the number of unknown pixels in the image, rather than the image resolution.
Hence, in a professional production setting where the unknown-opacity regions are typically narrower than the academic benchmarks, our algorithm has lower computational requirements.

\section{Spectral analysis}
\label{sec:spectral}

\begin{figure*}
{
\footnotesize
\begin{tabular}
{   K{0.143\linewidth}
    K{0.143\linewidth}
    K{0.143\linewidth}
    K{0.143\linewidth}
    K{0.143\linewidth}
    K{0.143\linewidth}
}
Input&
Only CM&
Only intra-\unt&
Only local&
CM \& intra-\unt&
CM, intra-\unt\ \& local
\end{tabular}
}
\showimagew[\linewidth]{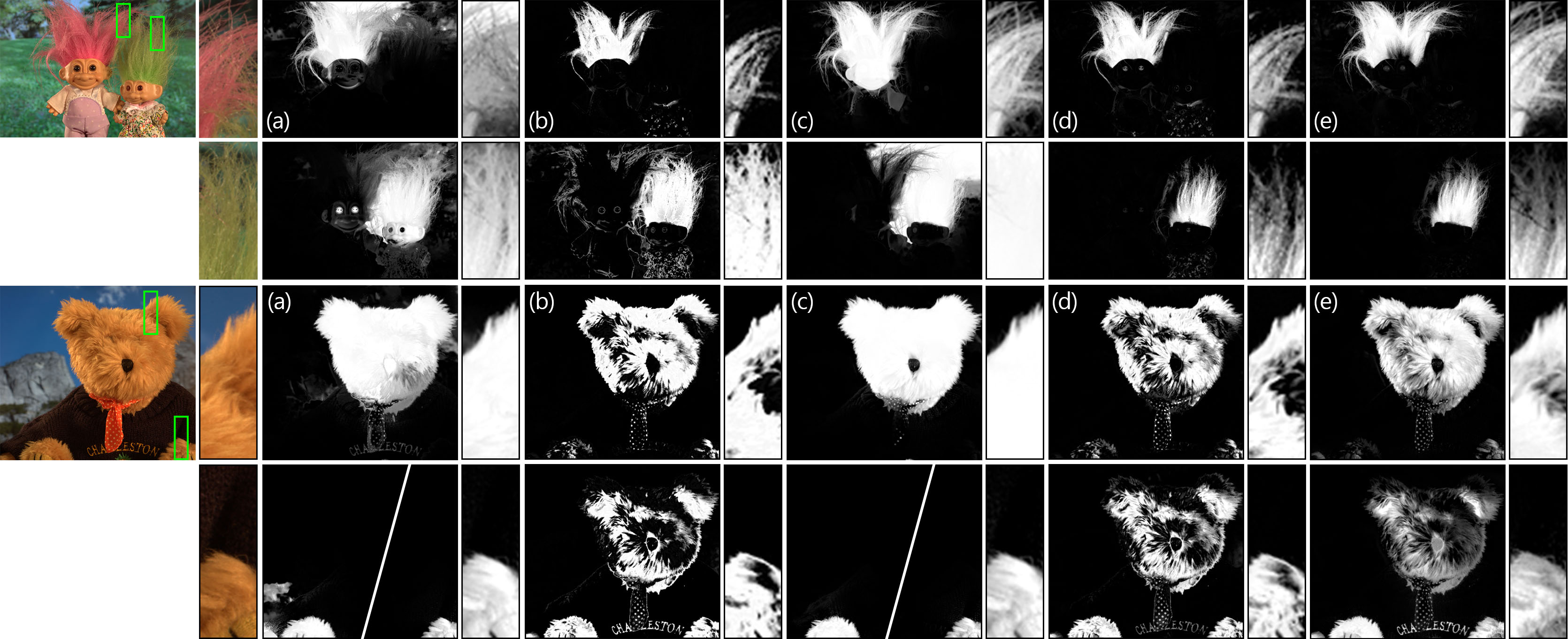}
\caption{
Selected matting components \cite{spectralpami} computed from Laplacian matrices constructed using different subsets of information flow.
Two components are included in the bottom examples for \emph{only CM} and \emph{only local} cases as the included parts appeared in separate components. 
}
\negvspace
\negvspace
\label{fig:spectral23}
\end{figure*}

The spectral clusters formed by Laplacians of affinity matrices can be effectively used to reveal characteristics of the constructed graph structure.
For instance, Levin~\etal\cite{closedformpami} analyze the matting affinity by looking at eigenvectors corresponding to the smallest eigenvalues of the matting Laplacian.
Spectral matting~\cite{spectralpami} uses the eigenvectors together with a sparsity prior to create a set soft segments, or \emph{alpha components}, that represent compact clusters of eigenvectors and add up to one for each pixel.
The alpha components provide a more distilled and clear visualization to analyze the affinity matrix.
In this section, we use the matting components computed using different subsets of information flows we defined for matte estimation to reveal the contribution of different flows at a higher level.

We compute the alpha components shown in Figure~\ref{fig:spectral23} using the public source code by Levin~\etal\cite{spectralpami}.
We exclude the \knt-to-\unt\ flow, which is only defined for the unknown regions as it requires explicitly defined known regions.
The resulting Laplacian matrix does not give meaningful spectral clustering because of the pixels with missing connections.
We overcome this issue for intra-\unt\ flow by defining it for the entire image instead of only the unknown region.
In our matting formulation, we use the color-mixture flow to create the main source of information flow between close-by similarly-colored pixels.
This approach creates densely connected graphs as both spatial and color distances are well accounted for in the neighborhood selection.
We observed that spectral matting may fail to create as many components as requested (10 in our experiments) in some images, as many regions are heavily interconnected.
Using the weighted average of neighboring colors for the flow
creates soft transitions between regions.

The intra-\unt\ flow connects pixels that have similar colors, with very little emphasis on the spatial distance.
This creates a color-based segmentation of the pixels, but as we compute the weights based on the feature distances, it is not typically able to create soft transitions between regions.
Rather, it creates components with alpha values at zero or one, or flat alpha regions with alpha values near $0.5$.

The local information flow, used as the only form of flow in the original spectral matting, creates locally connected components with soft transitions.

We observed a harmonious combination of positive aspects of these affinity matrices as they are put together to create our graph structure.
This provides a neat confirmation of our findings in the evaluation of our algorithm.
We analyze the characteristics of each flow more in detail through visual examples in the remainder of this section.

The top example in Figure~\ref{fig:spectral23} shows an input image with the matting components that include the green and the pink hair.
Color-mixture affinities give components that demonstrate the color similarity and soft transitions, but they typically bleed out of the confined regions of specific colors due to the densely connected nature of the graph formed by corresponding neighborhoods.
We clearly see the emphasis on color similarity for intra-\unt\ flow.
While the color clusters are apparent, one can easily observe that unrelated pixels get mixed into the clusters especially around transition regions between other colors.
We see a significant improvement already when these two flows are combined.
When the local information flow is added, which gives spatially confined clusters of many colors when used individually, we see smooth clusters of homogeneous colors.
The intricate transitions that were missed in the lack of the local flow are successfully captured when all three flows are included in the Laplacian definition.

The spatial connectivity versus color similarity characteristics are even more clearly observable in the bottom example of Figure~\ref{fig:spectral23}.
We see that bright and dark brown of the fur is clearly separated by intra-\unt\ flow in this example.
In contrast, color-mixture and local flows separate the fur into three spatial clusters and the sweater into two separate clusters despite the uniform color.
The combination, however, is able to successfully separate the dark and bright brown of the fur with smooth transitions.

The full Laplacian matrix we propose in this work blends the nonlocality of colors and spatial smoothness naturally.
This is the key characteristic of the proposed matting method.
When combined with \knt-to-\unt\ flow which addresses remote regions and holes inside the foreground, the proposed algorithm is able to achieve high performance in a variety of images as 
analyzed in Section~\ref{sec:results}.

\section{Sampling-based methods and \knt-to-\unt\ flow}
\label{sec:sampling}

\begin{table*}[t]
\caption{
SAD scores of top sampling-based methods on the matting benchmark against the \knt-to-\unt\ flow as a sampling based method, regularized by~\cite{shared}.
Blue shows the best performance among the methods listed here for each image-trimap pair.
Red marks the failure cases for the \knt-to-\unt\ flow.
}
\resizebox{\linewidth}{!}
{
{\setlength{\extrarowheight}{4pt}
\setlength\arrayrulewidth{1pt}
\rowcolors{2}{gray!25}{}
\begin{tabular}{
     !{\color{gray!25}\vrule} 
     l 
     !{\color{gray!25}\vrule} 
     c c c 
     !{\color{gray!25}\vrule} 
     c c c 
     !{\color{gray!25}\vrule} 
     c c c 
     !{\color{gray!25}\vrule} 
     c c c 
     !{\color{gray!25}\vrule} 
     c c c 
     !{\color{gray!25}\vrule} 
     c c c 
     !{\color{gray!25}\vrule} 
     c c c 
     !{\color{gray!25}\vrule} 
     c c c 
     !{\color{gray!25}\vrule}
     }
    \arrayrulecolor{gray!25}\hline
    &
    \multicolumn{3}{c !{\color{gray!25}\vrule} }{Troll} &
    \multicolumn{3}{c !{\color{gray!25}\vrule} }{Doll} &
    \multicolumn{3}{c !{\color{gray!25}\vrule} }{Donkey} &
    \multicolumn{3}{c !{\color{gray!25}\vrule} }{Elephant} &
    \multicolumn{3}{c !{\color{gray!25}\vrule} }{Plant} &
    \multicolumn{3}{c !{\color{gray!25}\vrule} }{Pineapple} &
    \multicolumn{3}{c !{\color{gray!25}\vrule} }{Plastic bag} &
    \multicolumn{3}{c !{\color{gray!25}\vrule} }{Net}
    \\
    & 
    \emph{S} &     \emph{L} &     \emph{U} &
    \emph{S} &     \emph{L} &     \emph{U} &
    \emph{S} &     \emph{L} &     \emph{U} &
    \emph{S} &     \emph{L} &     \emph{U} &
    \emph{S} &     \emph{L} &     \emph{U} &
    \emph{S} &     \emph{L} &     \emph{U} &
    \emph{S} &     \emph{L} &     \emph{U} &
    \emph{S} &     \emph{L} &     \emph{U}
    \\
    \arrayrulecolor{gray!25}\hline
    CSC \cite{csc} & 
    13.6 & 15.6 & \topscore{14.5} & 
    6.2 & 7.5 & 8.1 & 
    4.6 & 4.8 & 4.2 & 
    1.8 & 2.7 & 2.5 & 
    5.5 & 7.3 & 9.7 & 
    4.6 & 7.6 & 6.9 & 
    \topscore{23.7} & 23.0 & \topscore{21.0} & 
    26.3 & 27.2 & 25.2
    \\
    Sparse coding \cite{sparsecodestip} &  
    {12.6}  & {20.5}  & {14.8}  & 
    {5.7}  & \topscore{7.3}  & \topscore{6.4}  & 
    {4.5}  & {5.3}  & \topscore{3.7}  & 
    {1.4}  & {3.3}  & {2.3}  & 
    {6.3}  & {7.9}  & {11.1}  & 
    {4.2}  & {8.3}  & {6.4}  & 
    {28.7}  & {31.3}  & {27.1}  & 
    {23.6}  & {25.1}  & {27.3}
    \\
    KL-Div \cite{kldiv} &  
    {11.6}  & {17.5}  & {14.7}  & 
    \topscore{5.6}  & {8.5}  & {8.0}  & 
    {4.9}  & {5.3}  & {3.7}  & 
    {1.5}  & {3.5}  & {2.1}  & 
    {5.8}  & {8.3}  & {14.1}  & 
    {5.6}  & {9.3}  & {8.0}  & 
    {24.6}  & {27.7}  & {28.9}  & 
    \topscore{20.7}  & \topscore{22.7}  & \topscore{23.9}
    \\
    \knt-to-\unt\ inf. flow & 
    {12.0}  & \topscore{13.1}  & {14.6}  & 
    {7.5}  & {9.1}  & {8.9}  & 
    \topscore{3.9}  & \topscore{4.3}  & {3.8}  & 
    \topscore{1.4}  & \topscore{2.0}  & \topscore{2.0}  & 
    \topscore{5.3}  & \topscore{5.9}  & \topscore{8.0}  & 
    \topscore{2.7}  & \topscore{3.6}  & \topscore{3.3}  & 
    \failcase{37.2}  & \failcase{39.1}  & \failcase{35.8}  & 
    \failcase{47.2}  & \failcase{56.0}  & \failcase{41.9}
    \\
    Comp. Samp. \cite{comprehensive} &  
    \topscore{11.2}  & {18.5}  & {14.8}  & 
    {6.5}  & {9.5}  & {8.9}  & 
    {4.5}  & {4.9}  & {4.1}  & 
    {1.7}  & {3.1}  & {2.3}  & 
    {5.4}  & {9.8}  & {13.4}  & 
    {5.5}  & {11.5}  & {7.4}  & 
    {23.9}  & \topscore{22.0}  & {22.8}  & 
    {23.8}  & {28.0}  & {28.1}
    \\
    \arrayrulecolor{gray!25}\hline
\end{tabular}
}
}
\label{tab:benchmarkKtoU}
\end{table*}

The \knt-to-\unt\ flow introduced in Section~\ref{sec:knownToUnknown} connects every pixel in the unknown region directly to several pixels in both foreground and background.
While the amount of flow from each neighbor is individually defined by the computed color-mixture weights, we simplify the formulation and increase the sparsity of our linear system using some algebraic manipulations.
These manipulations, in the end, give us the weights $w^\fgm_p$ that go into the final energy formulation.

These weights, which show the connection of the unknown pixel to the foreground, are essentially an early estimation of the matte.
This estimation is done by individually selecting a set of neighbors for each pixel and computing an alpha based on the neighbor colors.
While our approach is fundamentally defining affinities, it has parallels with sampling-based approaches in natural matting~\cite{comprehensive,kldiv,csc,sparsecodestip}, which also select samples from foreground and background and estimates alpha values based on sample colors.
We compute confidence values for $w^\fgm_p$ that depends on the similarity of colors of neighbors from the foreground and background.
Sampling-based approaches also define confidence values for their initial estimation, typically defined by the \emph{compositing error}, $\|\ve{c} - (\alpha \ve{f} - (1-\alpha) \ve{b})\|^2$.

Conceptually, there are several fundamental differences between our computation of \knt-to-\unt\ flow and common strategy followed by sampling-based methods.
The major difference is how the samples are collected.
Sampling-based methods first determine a set of samples collected from known-alpha regions and do a selection for unknown pixels from this predetermined set using a set of heuristics. 
We, on the other hand, select neighbors for each unknown pixel individually via a k nearest neighbors search in the whole known region.
Using the samples, state-of-the-art methods typically use the compositing equation to estimate the alpha value from only one sample pair (a notable exception is CSC matting~\cite{csc}), while we use 14 samples in total to estimate the alpha by solving the overconstrained system using the method by Roweis and Saul~\cite{lle}.
These differences also change the computation time.
\knt-to-\unt\ flow can be computed in several seconds, while sampling-based algorithms typically take several minutes per image due to sampling and sample pair selection steps.


In order to compare the performance of \knt-to-\unt\ flow \emph{as} a sampling-based method in a neutral setting, in this experiment, we post-process $w^\fgm_p$ and our confidence values using the common regularization step~\cite{shared} utilized by top-performing sampling-based methods in the benchmark.
The quantitative results can be seen in Table~\ref{tab:benchmarkKtoU}.

As discussed in Section~\ref{sec:knownToUnknown}, \knt-to-\unt\ flow fails in the case of a highly-transparent matte (net and plastic bag examples).
This is due to the failure to find representative neighbors using the k nearest neighbor search.
Sampling-based methods are more successful in these cases due to their use of compositing error in the sample selection.
However, in the other examples, \knt-to-\unt\ flow appears as the top-performing method among the sampling-based methods in 12 of 18 image-trimap pairs and gives comparable errors in the rest.

The performance of our affinity-inspired approach against the state-of-the-art~\cite{comprehensive,kldiv,csc,sparsecodestip} gives us some pointers for a next-generation sampling-based matting method.
While one can argue that the sampling algorithms have reached enough sophistication, selection of a single pair of samples for each unknown pixel seems to be a limiting factor.
Methods that address the successful and efficient selection of \emph{many} samples for each unknown pixel will be more likely to surpass state-of-the-art performance.
Furthermore, determining the alpha values using more robust weight estimation formulations such as \refeq{eq:lle} instead of the more simple compositing equation~\refeq{eq:compositing} will likely improve the result quality.

\section{Limitations}
\label{sec:limitations}

As discussed in corresponding sections, the \knt-to-\unt\ flow does not perform well in the case of highly-transparent mattes.
We solve this issue via a simple classifier to detect highly-transparent mattes before alpha estimation.
However, this does not solve the issue for foreground images that partially have transparent regions.
For such cases, a locally changing set of parameters could be the solution.

\begin{figure}
\centering{
\showimagew[0.95\linewidth]{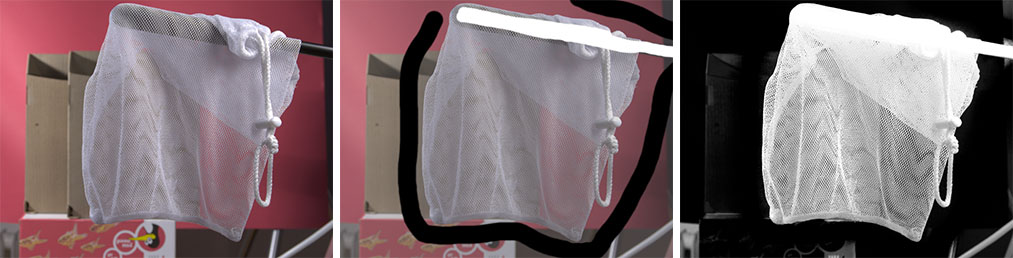}
}
\caption{
Our method fails gracefully in the case of sparse trimaps.
}
\negvspace
\negvspace
\label{fig:scribbleInput}
\end{figure}

The proposed matte estimation algorithm assumes dense trimaps as input.
In the case of sparse trimaps, generally referred as scribble input, our method may fail to achieve its original performance, as seen in Figure~\ref{fig:scribbleInput}.
This performance drop is mainly due to the \knt-to-\unt\ flow, which fails to find good neighbors in limited known regions, and intra-\unt\ flow which propagates alpha information based solely on color to spatially far away pixels inside the unknown region.

\section{Conclusion}
\label{sec:conclusion}

We proposed a purely affinity-based natural image matting method.
We introduced color-mixture flow, a specifically tailored form of LLE weights for natural image matting.
By carefully designing flow of information from the known region to the unknown region, as well as distributing the information inside the unknown region, we addressed several challenges that are common in natural matting.
We showed that the linear system we formulate outperforms the state-of-the-art in the alpha matting benchmark.
The characteristic contributions of each form of information flow were discussed through spectral analysis.
We extended our formulation to matte regularization and layer color estimation and demonstrate their performance improvements over the state-of-the-art.
We demonstrated that the proposed matting and color estimation methods achieve state-of-the-art performance in green-screen keying.
We also commented on several shortcomings of the state-of-the-art sampling-based methods by comparing them to our known-to-unknown information flow.



{\small
\bibliographystyle{ieee}
\bibliography{matting}
}

\end{document}